# Is semantic SLAM ready for embedded systems? A comparative survey

Calvin Galagain [a,b,*], Martyna Poreba [a], François Goulette [b]

[a] Université Paris-Saclay, CEA, List, F-91120, Palaiseau, France
[b] U2IS, ENSTA, Institut Polytechnique de Paris, 91120, Palaiseau, France

ABSTRACT

Semantic SLAM holds promise for robust robot navigation in complex environments, but its practicality on embedded systems remains uncertain. This paper presents a comparative survey of existing semantic visual SLAM systems from an embedded-deployment perspective. Unlike recent surveys centered on implicit representations, NeRFs, or 3D Gaussian Splatting as scene representations, our scope is the deployability of complete Semantic SLAM pipelines on resource-constrained robotics platforms, with particular emphasis on the NVIDIA Jetson AGX Orin. Three semantic-aware architectural approaches, namely Geometric SLAM, NeRF, and Gaussian Splatting-based SLAM, are evaluated with respect to accuracy, semantic reconstruction quality, memory footprint, power consumption, and throughput. Our Jetson measurements show that Semantic Geometric SLAM currently offers the most practical balance between accuracy and efficiency, whereas Gaussian Splatting-based systems require more than 16 GB of memory, draw more than 15 W, and remain far from real-time execution. The results also show that offline semantic preprocessing can mask the true cost of semantic integration. We therefore identify a research gap in adapting Semantic SLAM algorithms for embedded systems and propose directions for future work, including hardware-algorithm co-design and energy-efficient optimization.

## 1. Introduction

Simultaneous Localization and Mapping (SLAM) is a cornerstone technology in robotics and computer vision that enables autonomous systems to navigate and understand their environments. Over the past decade, the field has evolved significantly, moving from purely geometric approaches to methods enriched with semantic information. Initially, traditional SLAM techniques focused primarily on geometric representations, where the environment was modeled using geometric primitives such as points, lines, and surfaces. These methods, while effective at providing accurate localization and map building, often lacked contextual understanding of the environment, limiting their utility in complex or dynamic real-world scenarios. Semantic SLAM integrates high-level semantic information into the SLAM process, enhancing both map representation and localization performance. By incorporating object recognition, scene understanding, and semantic annotations, Semantic SLAM can identify and classify objects in the environment (for example, roads, cars, or pedestrians) and use this information to improve the robot's understanding of its surroundings. The integration of Artificial Intelligence (AI), particularly deep learning techniques, has been a driving force in advancing the development of Semantic SLAM. Convolutional Neural Networks (CNNs) and other neural networks are employed to perform tasks such as object detection, scene segmentation, and place recognition in real time. Neural Radiance Fields (NeRFs) have further revolutionized the domain by using neural networks to model the density and color of 3D spaces, allowing highly detailed reconstructions and novel view synthesis. More recently, 3D Gaussian Splatting (3DGS) has emerged as an efficient alternative, leveraging compact Gaussian primitives to represent scenes with continuous detail while reducing computational overhead. Together, these AI-driven techniques mark a transition toward richer, more versatile 3D representations that bridge geometry, semantics, and rendering efficiency.

Semantic SLAM has demonstrated the potential to improve localization accuracy, support dynamic environments, and enable robust map reconstruction. Fig. 1 presents a timeline of key advances in Semantic SLAM research from 2018 to 2026, highlighting the growing interest and accelerated development of these systems. Despite these advances, the increasing complexity of semantic extraction, dense scene representations, and online map optimization raises an important question: can complete Semantic SLAM pipelines operate under the memory, power, portability, and real-time constraints of embedded robotic platforms?

* Corresponding author at: Université Paris-Saclay, CEA, List, F-91120, Palaiseau, France.
*E-mail address:* calvin.galagain@cea.fr (C. Galagain).

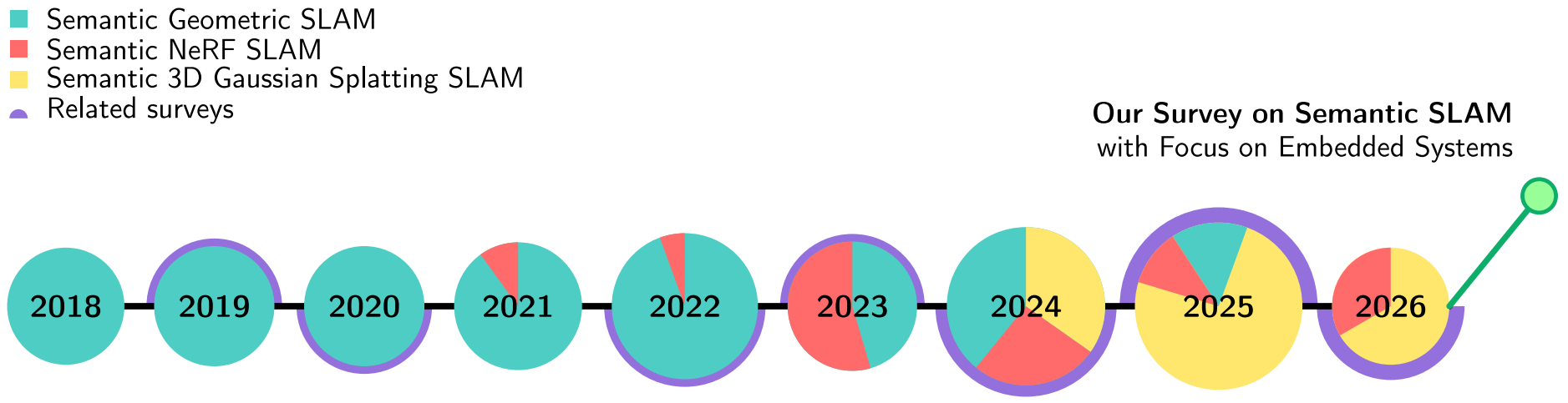

**Fig. 1.** Timeline of representative Semantic SLAM developments from 2018 through August 2026. The progression is categorized into three primary approaches: Semantic Geometric SLAM (Turquoise), Semantic NeRF SLAM (Coral), and Semantic 3DGS SLAM (Yellow). Circle sizes qualitatively indicate the relative publication volume among the works considered in this survey. Purple background semicircles encode the number of related surveys assigned to each year using the final publication year when available; their visible radial margin increases with the survey count. The present embedded-focused survey is shown separately in Green, to the right of the surveyed literature.

**Table 1**
Comparison of existing surveys directly related to Semantic SLAM and neural scene representations. Y indicates a primary focus, P indicates partial coverage, and – indicates that the topic is not systematically addressed. HW eval. denotes comparative measurements performed on a common embedded platform. Sem. cost indicates whether the computational cost of online semantic inference is explicitly distinguished from that of offline semantic preprocessing.

| Survey | SemSLAM | NeRF | 3DGS | Emb. | HW eval. | Sem. cost |
|---|---|---|---|---|---|---|
| Xia et al. [1] | Y | – | – | – | – | – |
| Chen et al. [2] | Y | P | – | – | – | – |
| Chen et al. [3] | Y | – | – | – | – | – |
| Wang et al. [4] | Y | – | – | – | – | – |
| Kanso et al. [5] | Y | P | P | – | – | – |
| Miao et al. [6] | Y | P | P | P | – | – |
| Canh et al. [7] | Y | P | P | – | – | – |
| Yang et al. [8] | P | Y | – | – | – | – |
| Wang et al. [9] | P | Y | – | P | – | – |
| Irshad et al. [10] | P | Y | P | P | – | – |
| Tosi et al. [11] | P | Y | Y | P | – | – |
| Chen and Wang [12] | – | P | Y | – | – | – |
| Bao et al. [13] | P | P | Y | – | – | – |
| Wang et al. [14] | P | – | Y | P | – | – |
| Salhi et al. [15] | – | – | – | Y | – | – |
| Picard et al. [16] | – | P | – | Y | P | – |
| Tahir and Parasuraman [17] | – | – | – | Y | – | – |
| **Ours** | **Y** | **Y** | **Y** | **Y** | **Y** | **Y** |

### *1.1. Positioning with respect to existing surveys*

To provide an up-to-date and systematic positioning, we consider surveys published or made available online through 7 August 2026 that directly address Semantic SLAM, implicit or NeRF-based SLAM, 3D Gaussian Splatting, 3DGS-based SLAM, or embedded robotic perception and reconstruction. These reviews can be grouped into four complementary categories: general and Semantic SLAM surveys, implicit and NeRF-based SLAM surveys, 3DGS and 3DGS-SLAM surveys, and embedded localization, perception, and reconstruction surveys. Table 1 compares their coverage with that of the present work.

Earlier surveys examine the transition from geometric to Semantic SLAM, the extraction and association of semantic information, semantic map representations, and the handling of dynamic environments [1–4]. More recent reviews extend this analysis. Kanso et al. [5] provide a comprehensive overview of Semantic SLAM methods, applications, datasets, evaluation metrics, and reproducibility. Miao et al. [6] focus on open-world Semantic SLAM, including zero-shot open-vocabulary understanding, incremental semantic expansion, and multimodal fusion. Canh et al. [7] propose a modular formulation organized around visual localization, semantic feature extraction, mapping, data association, and loop-closure optimization. These surveys provide comprehensive algorithmic and application-oriented taxonomies, but do not evaluate representative complete pipelines on a common embedded platform using measured memory, power, and throughput.

Yang et al. [8] review implicit SLAM architectures with an emphasis on NeRF-based scene representations. Wang et al. [9] and Irshad et al. [10] examine NeRFs and neural fields more broadly in robotic perception, reconstruction, and interaction. Tosi et al. [11] analyze the evolution of SLAM from classical pipelines toward NeRF and 3DGS representations. Their primary objective is to organize the representation, tracking, mapping, and rendering landscape rather than to measure the end-to-end deployability of Semantic SLAM pipelines on resource-constrained hardware.

General 3DGS surveys examine Gaussian representations, optimization, rendering, compression, and applications [12,13]. More recently, Wang et al. [14] specifically review 3DGS-SLAM methods according to rendering quality, tracking accuracy, reconstruction speed, memory consumption, and robustness. Although these works discuss algorithmic efficiency and memory requirements, they do not jointly compare semantic geometric, NeRF-based, and 3DGS-based SLAM through measurements performed on a common embedded platform.

Salhi et al. [15] review multimodal localization from an embedded algorithm and architecture adequacy perspective. Picard et al. [16] examine real-time SLAM-based 3D reconstruction under embedded resource constraints. More broadly, Tahir and Parasuraman [17] review edge computing for latency-sensitive robotic applications. These works address deployment and hardware constraints, but they do not systematically study how semantic inference interacts with geometric, NeRF-based, and 3DGS-based SLAM architectures.

Beyond survey articles, several benchmarking efforts provide standardized evaluations of SLAM robustness, implicit RGB-D SLAM, NeRFs in robotics, and NeRF/3DGS reconstruction [18–22]. These studies mainly compare trajectory accuracy, robustness, or reconstruction quality. They do not evaluate complete Semantic SLAM pipelines under a common set of embedded memory, power, software portability, and online semantic-processing constraints.

Table 1 reveals a gap between algorithm-centered surveys of Semantic SLAM and neural scene representations, on the one hand, and system-centered surveys of embedded robotics, on the other hand. Existing Semantic SLAM surveys primarily classify semantic extraction, mapping, data association, and dynamic-scene handling. NeRF- and 3DGS-oriented surveys mainly examine scene representation, tracking, reconstruction, rendering quality, and algorithmic efficiency. Embedded surveys, in turn, address localization, reconstruction, or edge-computing constraints without systematically analyzing the cost of semantic integration. To the best of our knowledge, no previous survey jointly compares semantic geometric, NeRF-based, and 3DGS-based SLAM architectures under a common embedded-deployment perspective while reporting localization accuracy, semantic quality, memory footprint, power consumption, software portability, and online throughput. Existing comparisons also rarely distinguish complete online semantic inference from configurations relying on precomputed semantic masks.

### 1.2. Scope and contributions

Accordingly, the objective of this paper is not to provide another broad taxonomy of Semantic SLAM, NeRF, or 3DGS methods. Instead, we examine whether complete Semantic SLAM pipelines can satisfy the memory, energy, portability, and real-time requirements of embedded robotic platforms. The study covers three fundamentally different architectural families: semantic geometric SLAM, semantic NeRF-based SLAM, and semantic 3DGS-based SLAM. The main contributions of this work are as follows:

- We introduce a comparative survey of Semantic SLAM systems from an embedded-deployment perspective, offering an in-depth analysis of how semantic modules are incorporated into the SLAM pipeline and how these choices affect runtime, memory, and energy requirements.
- We conduct a study of selected SLAM systems, which are characterized by fundamentally different architectures, including RDS-SLAM [23], VDO-SLAM [24], Dynamic-VINS [25], GS3LAM [26], SNI-SLAM [27] and SGS-SLAM [28]. This analysis combines accuracy and semantic-quality comparisons with a resource-utilization assessment on the NVIDIA Jetson AGX Orin, focusing on memory usage, power consumption, throughput, and the cost of semantic processing.
- We derive hardware-aware conclusions for embedded Semantic SLAM, showing how deployment feasibility changes when offline semantic preprocessing, ARM/CUDA portability, memory footprint, and power consumption are considered alongside ATE and mIoU.

## 2. State of the art

SLAM systems operate on several key assumptions. First, they presume that the environment is largely static, treating any dynamic objects as noise or outliers. Second, they rely on the precision of the sensors to ensure accurate geometric measurements. Although these assumptions work well for structured and predictable settings, they limit the applicability of SLAM in environments with significant dynamic elements or complex object interactions. Semantic SLAM enhances traditional SLAM systems by incorporating high-level semantic insights. This enables the extraction of information about objects' functional attributes, their interactions with surrounding elements, and the overall context of the environment. Semantic SLAM unlocks new opportunities for smarter and more adaptable robotic behavior in complex, real-world settings. Enhanced localization is achieved by leveraging contextual cues from segmented scenes, allowing systems to anchor themselves more accurately in their environment. In dynamic settings, semantic understanding facilitates the exclusion of transient objects, refining the consistency and reliability of localization. Furthermore, semantic mapping enables richer reconstructions, either through sparse keypoints or dense representations such as NeRF and 3D Gaussian Splatting. These capabilities open new frontiers for robotic applications, particularly in domains that require both precision and adaptability. However, **this raises a critical question: at what cost?**

In this section, the integration of semantic information in SLAM systems is explored. This is typically achieved through the use of neural networks, which can be incorporated into SLAM pipelines in various ways. These networks provide robust scene understanding by segmenting images, detecting objects, or generating dense semantic representations. We analyze Semantic-aware Geometric SLAM approaches 2.1 where semantics serve a dual purpose: managing scene dynamics by identifying and excluding moving objects, and enhancing map representations, regardless of whether they involve sparse structures like point clouds or volumetric frameworks such as OctoMap [29]. Furthermore, we examine advanced algorithms that focus on improving the quality of the reconstructed scene, including the NeRF 2.2 or 3DGS 2.3 methods, which utilize neural representations for the generation of detailed and photorealistic maps. Table 2 summarizes the methods reviewed in this survey, highlighting critical parameters, including the type of tracker chosen as a baseline, the neural network used to deliver semantic information and the hardware platform on which each system was tested. These platforms range from high-performance GPUs in desktop environments to limited resources embedded systems.

### 2.1. Geometric SLAM meets semantics

Geometric SLAM represents the scene using explicit geometric features such as points, lines, or planes. It constructs maps by directly detecting and matching these features across camera frames. Pose estimation in Geometric SLAM is computed using traditional geometry-based methods, such as Perspective-n-Point (PnP) or Iterative Closest Point (ICP), which rely on clear feature correspondences. Thus, the typical architecture of a Geometric SLAM consists of three main components: tracking to estimate pose, mapping to reconstruct the environment, and optimization through bundle adjustment with loop closure detection, as illustrated in Fig. 2. The main idea of Semantic Geometric SLAM methods is to enhance traditional SLAM by incorporating semantic information, such as object recognition and classification, alongside geometric data. This semantic information is typically introduced through the integration of deep learning models or pre-trained classifiers, which process sensor data to identify and label objects in the environment. For example, object detection models such as Faster R-CNN [30], and the YOLO family of models [31] including its various iterations like YOLOv3, YOLOv5, YOLOv8, or, more recently, foundation models such as DINOv2 [32], are commonly used to identify and locate objects. For semantic segmentation, various networks DuNet [33], PSPNet [34], SegNet [35], FCHarDNet [36], or BiSeNetv2 [37] classify each pixel into meaningful categories, providing a comprehensive understanding of the elements of the scene. In contrast, instance segmentation methods such as Mask R-CNN [38], YOLACT [39], and more recent approaches, including the Segment Anything Model (SAM) [40], enhance this capability by combining object detection with pixel-level delineation, enabling the precise identification and separation of individual objects within a scene.

Some approaches [41–43] leverage the integration of panoptic segmentation through CNN models, such as PanopticFPN [44], or use promptable and interactive models for segmenting-everything-everywhere, like SEEM [45]. It enhances detection, segmentation, and instance segmentation capabilities by providing a unified framework that not only classifies objects but also differentiates individual object instances and handles background regions. Semantic management in SLAM systems can be seen to closely align with the state-of-the-art techniques that were prevalent at the time. Early Semantic SLAM approaches, developed between 2018 and 2021, predominantly integrated instance segmentation using Mask R-CNN, which is known for its ability to provide high-quality segmentation and object detection. However, Mask R-CNN is computationally intensive, making it less suitable for real-time applications. In contrast, more recent Geometric SLAM methods have shifted toward incorporating semantics through object detectors. It can be observed that recent systems tend to integrate foundation models such as DinoV2 or SAM. This trend arises because foundation models offer significant advantages in terms of generalization and adaptability. By leveraging large-scale pre-trained models systems can benefit from robust feature extraction and improved performance across a wide range of tasks without requiring task-specific training data.

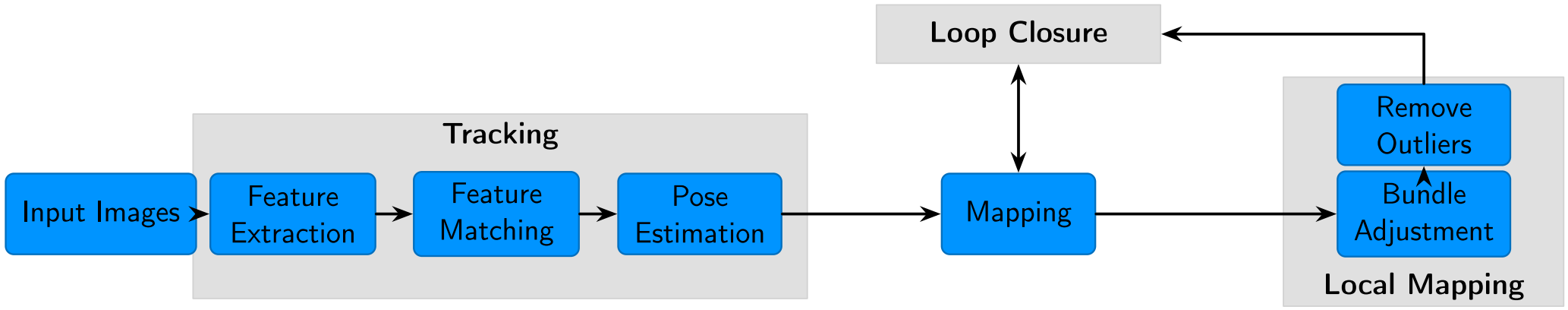

**Fig. 2.** Typical pipeline of a feature-based geometric SLAM system. Tracking extracts and matches visual features for pose estimation; mapping reconstructs the environment; local bundle adjustment and outlier rejection refine the map; and loop closure feeds global corrections back into mapping.

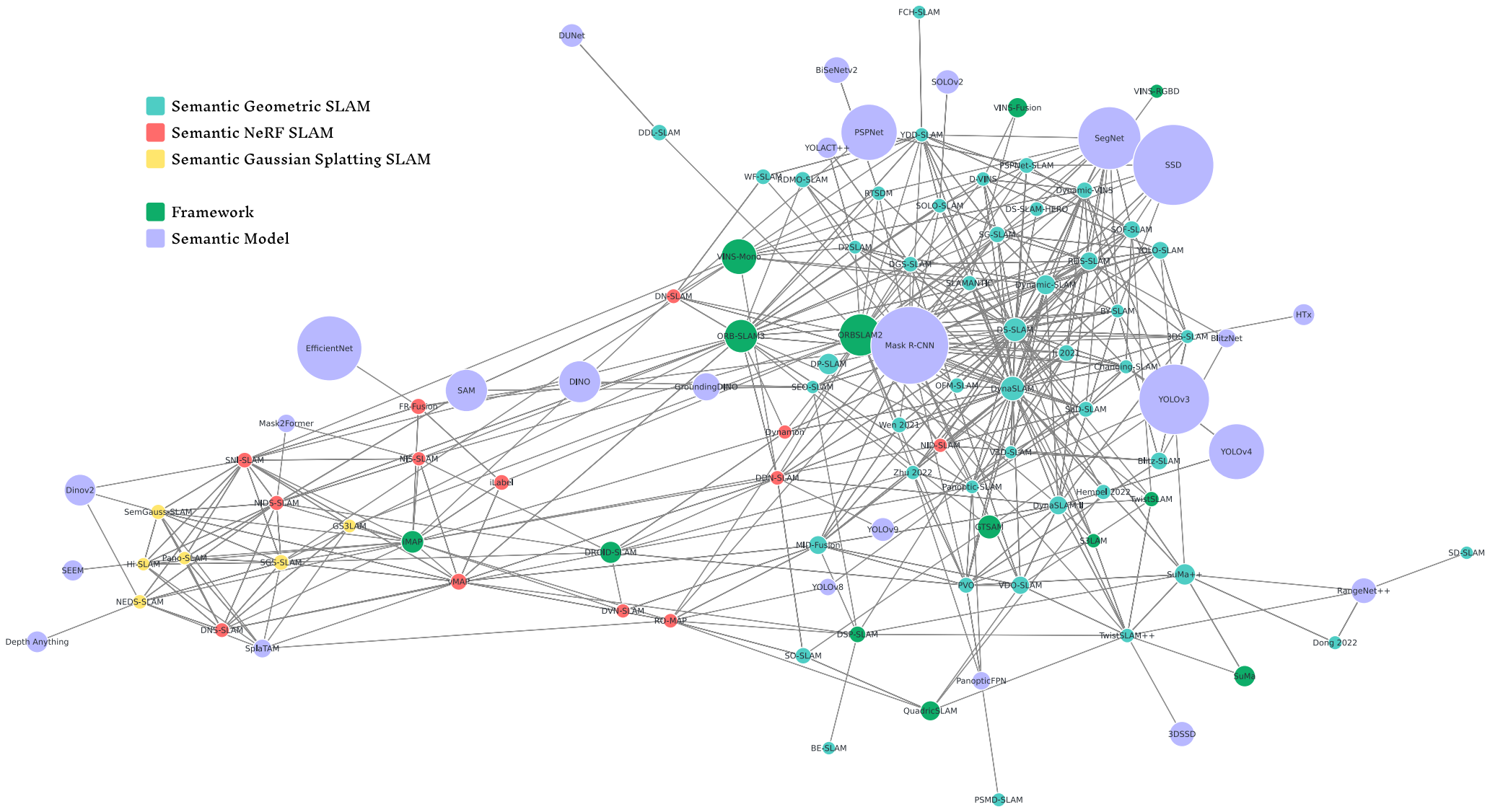

**Fig. 3.** Citation network and relationships in Semantic SLAM research. The visualization represents five distinct categories: Geometric SLAM approaches (Turquoise), NeRF-based SLAM methods (Coral), GS-based SLAM techniques (Yellow), Framework implementations (Green), and Segmentation/Detection Models (Purple). Node sizes correspond to citation impact, with larger nodes indicating higher citation counts. The network topology reveals two distinct clusters: (1) A left-oriented cluster dominated by NeRF and Gaussian Splatting SLAM approaches, sharing common theoretical foundations and citation patterns and (2) A right-oriented cluster primarily composed of geometric approaches, centered around foundational works like ORB-SLAM.

In Semantic or Object-aware SLAM systems, the base SLAM framework generally maintains its original structure. Adding semantic layers requires some modifications, such as introducing new processing pipelines, but rarely necessitates a complete overhaul of the system. As shown in Table 2 and Fig. 3, the majority of Semantic Geometric SLAM systems available to date are based on extended versions of ORB-SLAM, specifically ORB-SLAM2 [46] and ORB-SLAM3 [47]. These systems have become widely adopted due to their robustness and versatility, handling a variety of scenarios such as monocular, stereo, and RGB-D setups, making them an ideal starting point for integrating semantic information. There are two primary approaches to incorporating online semantics into SLAM systems: one involves applying semantics selectively to keyframes (Fig. 4), while the other integrates semantic information into every frame. In the first approach, as seen in [48], RTSDM [49], RDS-SLAM [23], SOLO-SLAM [50], RDMO-SLAM [51], or D-VINS [52], semantic information is extracted and applied only to keyframes, selected based on significant pose changes, scene content, tracking quality, or optimization needs. This strategy reduces the computational burden by limiting the frequency of semantic processing, making it more suitable for real-time applications on platforms with limited resources. In contrast, the second approach involves integrating semantic information into every single frame, which provides a more continuous and dynamic understanding of the environment, but at the cost of increased computational requirements. This approach is typically used in systems that prioritize precision and can afford to process more frames, such as [53], FCH-SLAM [54], WF-SLAM [55], SLAMANTIC [56], SaD-SLAM [57], DS-SLAM [58], Dyna-SLAM [59], SOF-SLAM [60], PSPNet-SLAM [61], OFM-SLAM [62], DDL-SLAM [63], D2SLAM [64], YOEC-DSLAM [65], SEO-SLAM [66], or V3D-SLAM [67]. These systems work well in scenarios where the environment is highly dynamic, but the increased processing cost makes them less suitable for hardware platforms with limited resources. VDO-SLAM [24] is not assigned to either online-inference category here because, in the evaluated configuration, its instance masks are generated through offline Mask R-CNN preprocessing before the tracking and mapping loop.

Once semantic information is obtained, distinguishing between static and mobile objects becomes a crucial step in improving the performance of Semantic SLAM. Various approaches have been developed to address this challenge, ranging from simple assumptions to advanced learning methods. One straightforward method assumes that potentially mobile objects, such as vehicles and pedestrians, are in motion and should be filtered out. Although effective in certain scenarios, this assumption may not always be valid, especially for stationary objects that are inherently capable of mobility. To address such limitations, more robust methods have been proposed to accurately determine dynamic objects using additional cues. For example, geometric clustering [48] or HDBSCAN clustering (Hierarchical Density-Based Spatial Clustering) [68] serve to analyze spatial patterns to separate static and dynamic features. Epipolar constraints are also used by matching points in the environment and classifying them according to their behavior within potential dynamic regions [55,64,69,70]. DS-SLAM [58] computes an optical flow pyramid. In [60], a novel approach for detecting dynamic features, termed semantic optical flow, is introduced. This method employs a tightly coupled framework that effectively leverages the dynamic characteristics of features embedded within semantic and geometric information. YOLO-SLAM [71] and the method in [72] combine object detection with geometric constraints between frames to filter dynamic objects in monocular images. Other techniques include multiview geometry [59,63,65], which takes advantage of multiple perspectives of the scene to identify inconsistencies in object locations, and PROSAC-based methods [61], which prioritize stable matching points in dynamic regions.

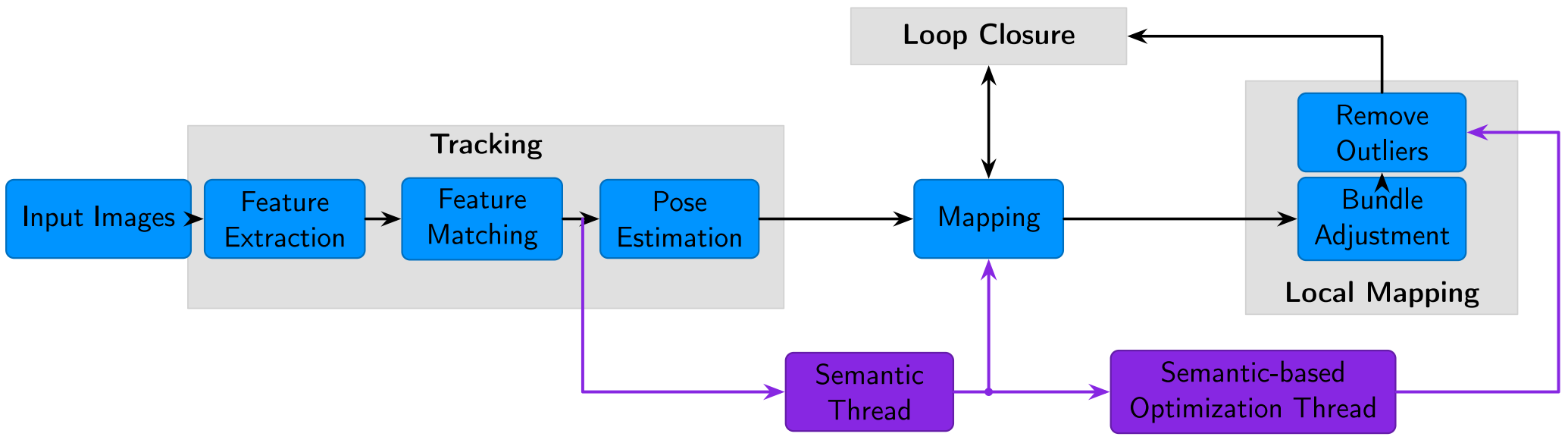

**Fig. 4.** Architecture of RDS-SLAM [23] based on ORB-SLAM3. The traditional geometric SLAM components are shown in blue, while the purple elements highlight the additional semantic threads. The Semantic Thread processes dynamic points detection, and the Semantic Optimization Thread enhances both mapping accuracy and localization through semantic information.

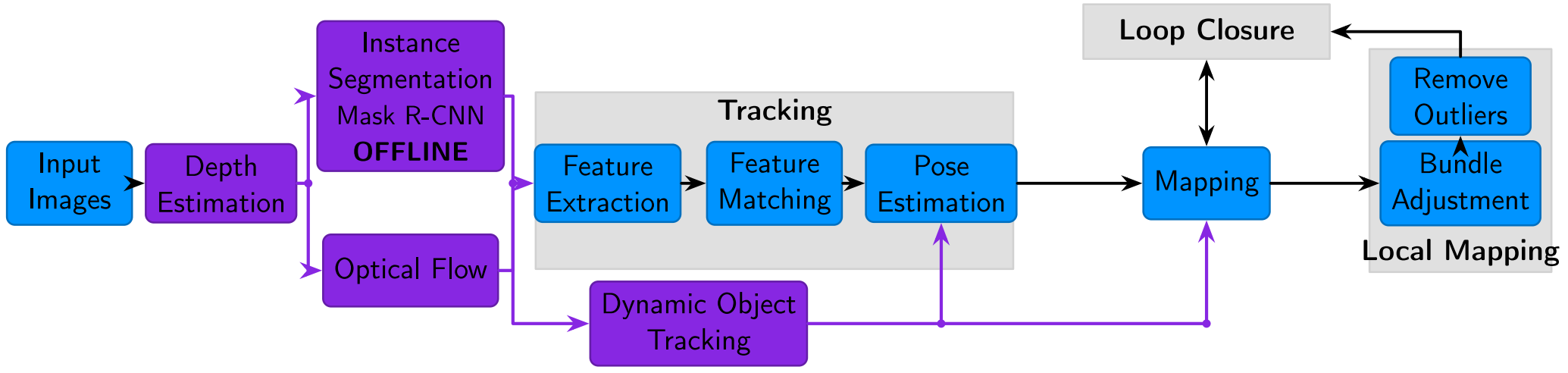

**Fig. 5.** Overview of the VDO-SLAM† architecture [24]. The system extends traditional geometric SLAM (in blue) with additional components (in purple) to handle dynamic environments. The pipeline introduces a preprocessing stage with depth estimation, instance segmentation, and optical flow computation. In the configuration evaluated in this survey, the instance masks are produced by Mask R-CNN offline and semantic inference is excluded from the reported online runtime; this distinction is denoted by †. A parallel dynamic object tracking stream complements the standard geometric tracking, enabling the system to simultaneously handle both static and dynamic elements in the mapping process.

In Figs. 1 and 3, a few representative and widely cited works can be distinguished. RDMO-SLAM, and RDS-SLAM are all built upon the ORB-SLAM3 pipeline and integrate Mask R-CNN for instance segmentation, which is applied exclusively to keyframes. In RDS-SLAM semantic information is stored within the multi-map representation framework, ATLAS [73], and serves to exclude outliers from tracking through a data association algorithm. The system introduces two novel threads, namely a semantic thread and a semantic optimization thread (highlighted in purple in Fig. 4), which run in parallel with the existing threads. This design enables efficient integration of semantic data without delaying the tracking process. RDMO-SLAM extends the capabilities of RDS-SLAM by implementing a semantic label prediction mechanism that utilizes optical flow to efficiently extract additional semantic information. By analyzing optical flow patterns, the system predicts the semantic attributes of keyframes using previously obtained semantic labels, minimizing dependence on resource-intensive direct segmentation processes.

Similarly, SOLO-SLAM [50] extends the ORB-SLAM3 framework by adding a dedicated semantic processing thread that leverages the SOLOv2 function package [74]. Semantic data are incorporated through instance segmentation applied to keyframes only. This semantic thread updates the dynamic probabilities and semantic attributes of the map points. To address the challenges of recognizing dynamic points, the system employs a quadratic filtering method based on dynamic degrees and geometric constraints.

In contrast, MID-Fusion [75] generates object-level dynamic maps by using an object-centric representation that integrates the geometric, semantic, and motion properties of objects within the scene. Notably, in MID-Fusion, Mask R-CNN segmentation serves as a preprocessing step. This step is executed on the GPU to detect and segment dynamic objects in the environment before the core SLAM process begins, ensuring that dynamic elements are identified and managed effectively from the outset.

On its side, OFM-SLAM [62] generates dynamic object masks for each frame by combining Mask R-CNN, which detects potential moving objects, with optical flow techniques to identify real dynamic feature points. This integration enables the system to accurately track and differentiate between static and dynamic elements in real-time environments, enhancing its robustness and adaptability. Furthermore, similar to MID-Fusion, instead of constructing a sparse semantic map, OFM-SLAM builds a semantic octree map. This map is constructed using the results of semantic segmentation combined with a dense point cloud map, providing a more detailed and structured representation of the environment.

In VDO-SLAM [24], Mask R-CNN is also used for semantic extraction and preprocessing steps (Fig. 5); however, unlike other systems, its tracking and mapping pipeline is not built on existing SLAM architectures. To distinguish between static and dynamic objects, VDO-SLAM uses PWC-Net [76], an advanced optical flow network, to estimate dense pixel motion between consecutive frames. However, a major limitation of this system is that it is only applicable to pre-recorded benchmarks, making it unsuitable for deployment in real-time, real-world scenarios.

Table 2 is intentionally selective rather than exhaustive. It preserves the principal methods from the original comparison and adds only historically influential systems, highly cited representatives of a distinct architecture, major recent peer-reviewed contributions, and methods directly relevant to embedded deployment. For papers published in 2025 and 2026 that have not yet had time to accumulate citations, venue, methodological novelty, implementation availability, and relevance to this survey are used instead. The table merges preprint and publisher versions and reports the final version-of-record year whenever one exists. Pure reconstruction or semantic-mapping methods requiring external, ground-truth, or COLMAP poses are excluded, as are LiDAR-only systems. The literature search and citation snapshot were frozen on 7 August 2026.

Recent Semantic Geometric SLAM approaches are still largely based on ORB-SLAM, but they increasingly integrate object detection. YOEC-DSLAM [65] focuses on improving SLAM in dynamic environments by integrating lightweight semantic segmentation with geometric filtering based on the YOLOv5s-SEG backbone, optimized for real-time performance. Key features include GhostConv modules, SimSPPF for efficient computation, and CBAMC3 modules for enhanced feature extraction. These components enable for accurate segmentation of dynamic objects with minimal computational cost. By combining segmentation with multi-view geometry, YOEC-DSLAM identifies and filters dynamic feature points to improve trajectory estimation and map quality.

3DS-SLAM [68] introduces a method for handling dynamic environments by incorporating 3D object detection with a Hybrid Transformer (HTx) architecture. Unlike methods that rely only on 2D semantic information or geometric constraints, 3DS-SLAM combines semantic and geometric cues in a tightly integrated framework. The HTx architecture enables real-time 3D object detection while maintaining robustness under dynamic conditions. Geometric filtering is performed using the HDBSCAN algorithm, which identifies objects with notable depth differences.

V3D-SLAM [67] segments the objects using YOLOv8 and combines RGB-D SLAM with a spatial Hough voting mechanism to detect moving objects. Object segmentation is refined with Chamfer distance to account for intra-object motion, and ORB-based feature extraction is applied to static regions. Together with pose graph optimization, this approach improves trajectory estimation and mapping in dynamic environments.

SEO-SLAM [66] enhances semantic SLAM by utilizing Vision-Language Models (VLMs) and Multimodal Large Language Models (MLLMs) for object-level semantic mapping in cluttered environments. It addresses challenges such as identifying similar objects, correcting landmark errors, and reducing detection biases using dynamic multi-class confusion matrices. MLLM feedback is used to refine labels and resolve perceptual aliasing, improving semantic accuracy and map consistency.

Finally, some authors use semantics in swarm control systems to improve precision and collaboration. $D^2$SLAM [64] presents a distributed and decentralized SLAM system for aerial robot swarms. It separates state estimation into near-field and far-field categories to achieve precise relative localization for nearby robots and consistent global trajectories for distant ones. The system includes $D^2$VINS, a distributed visual–inertial state estimator optimized using ADMM, and $D^2$PGO, a distributed pose graph optimizer based on the asynchronous ARock algorithm. It supports stereo and omnidirectional cameras and is designed to handle network delays and scale effectively, making it suitable for multi-robot applications.

### 2.2. Semantic NeRF SLAM

NeRFs, introduced by Mildenhall et al. [184] in 2020, have revolutionized 3D scene representation by modeling a continuous volumetric function through a neural network. NeRF maps 3D spatial coordinates and viewing directions to color and density, enabling photorealistic novel view synthesis with remarkable multi-view consistency. Unlike traditional discrete representations, such as voxel grids or meshes, NeRF offers a compact yet highly expressive approach, achieving state-of-the-art results in complex lighting and geometric scenarios.

Integrating NeRF into SLAM systems enhances autonomous navigation and robotic mapping by addressing limitations of Geometric SLAM. Although effective in many environments, traditional methods struggle with complex scenes, textureless surfaces, and occlusions, leading to challenges in generating accurate 3D reconstructions, particularly in scenarios with difficult lighting or repetitive patterns. NeRF-based SLAM tightly couples pose calculation and map construction within an optimization framework that integrates computer vision principles with neural networks. Although scene representation differs significantly between geometric and NeRF-based approaches, these optimization components show that the fundamental geometric SLAM principles are still applicable and adaptable to neural radiance field architectures. NeRF-based SLAM approaches that utilize all or part of the well-known ORB-SLAM are still numerous [130,132,133,135,137]. For others, the iMAP framework [126] serves as a foundational approach. iMAP initializes poses through frame-to-frame feature matching and refines them using global optimization. This enables joint optimization of camera poses and the neural radiance field by minimizing differences between observed images and those rendered from the NeRF model. The system iteratively improves both pose estimation and mapping. The iMAP system operates with two concurrent processes: tracking, which optimizes the pose of the current frame relative to the fixed network, and mapping, which simultaneously optimizes the poses of the network and selected keyframe. This approach allows for consistent updates to both the camera trajectory and the underlying scene representation. NIDS-SLAM [130] incorporates loop closure detection modules to optimize trajectory estimation when re-visiting previously mapped areas, effectively reducing accumulated drift. DNS-SLAM [131] and DDN-SLAM [137] implement Global Bundle Adjustment (Global BA), jointly optimizing camera poses and NeRF scene representation. Additionally, DDN-SLAM combines loop closure detection and Global BA in a unified framework, demonstrating how multiple geometric SLAM components can be effectively integrated.

Semantic NeRFs (Fig. 6) provide a method for dense Semantic SLAM by incorporating semantic features into NeRF frameworks along with geometry and appearance. These methods aim to improve scene understanding while preserving the benefits of continuous neural representations. Over time, these approaches have progressed from basic feature integration to more advanced multi-modal architectures, addressing various challenges in semantic mapping. The semantics can either be pre-processed and inserted as an input to the system [133,137,138], or handled through parallel processing threads, which operate on either keyframes [125,127,130–132,134,135,139] or every frame [27,136]. An early example is SNI-SLAM [27], which employs a three-part architecture to integrate geometry, appearance, and semantics. The system uses a cross-attention mechanism to fuse features, allowing bidirectional information flow between modalities. RGB-D input is processed through separate MLP networks for each modality and then combined using cross-attention modules. The hierarchical semantic representation incorporates multi-scale feature planes to capture both coarse and fine scene details. The decoder architecture facilitates one-way feature correlation, minimizing interference while enabling effective information sharing. Optimization involves photometric, geometric, and semantic losses, along with a feature loss designed to operate at a higher representation level.

**Table 2**
Semantic SLAM Systems Overview. Methods marked with † use semantic masks or labels precomputed offline; semantic inference is therefore not part of their online SLAM pipeline in the evaluated configurations. We categorize the different methods detailed in Sections 2.1 to 2.3. Reported from the second column on the left to the second column on the right are: the method name (a) and publication year (b), followed by the: (c) Input modalities they can process; (d) Foundational tracker system on which they are built; (e) Model used for object/semantic segmentation; (f) Way of integrating semantics which can be per frame or keyframe-only; (g) Presence of a moving consistency check to handle dynamic environments, eliminating the need to rely on assumptions about object dynamics; (h) Hardware used for testing the SLAM system. Finally, (i) we indicate whether the source code has been released.

| | | | | Semantic Module | | | | |
|---|---|---|---|---|---|---|---|---|
| (a) | (b) | (c) | (d) | (e) | (f) | (g) | (h) | (i) |
| Method | Year | Input | Framework | Model | Keyframe | Dynamic | Hardware | Code |
| **Semantic Geometric SLAM** [2.1] | | | | | | | | |
| SLAM++ [77] | 2013 | RGB-D | Dense object-level SLAM | Known 3D object models | | | | |
| SemanticFusion [78] | 2017 | RGB-D | ElasticFusion | CNN semantic segmentation | | | NVIDIA Titan X | • |
| DS-SLAM [58] | 2018 | RGB-D | ORB-SLAM2 | SegNet | | • | Intel i7 CPU, P4000 GPU | • |
| DynaSLAM [59] | 2018 | RGB-D | ORB-SLAM2 | Mask R-CNN | | • | | • |
| Detect-SLAM [79] | 2018 | RGB-D | ORB-SLAM2 | SSD | | • | | |
| MaskFusion [80] | 2018 | RGB-D | Surfel multiobject SLAM | Mask R-CNN, geometric segmentation | | • | | • |
| Fusion++ [81] | 2018 | RGB-D | Volumetric object-level SLAM | Mask R-CNN | | | GTX Titan X | • |
| MID-Fusion [75] | 2018 | RGB-D | | Mask R-CNN | | • | Intel Core i7-7700 CPU | • |
| SOF-SLAM [60] | 2019 | RGB-D | ORB-SLAM2 | SegNet | | • | | |
| SuMa++ [82] | 2019 | LiDAR | SuMa [83] | FCN RangeNet++ | | | Nvidia Quadro P4000 with 8 GB RAM. | • |
| SLAMANTIC [56] | 2019 | RGB-D | ORB-SLAM2 | Mask R-CNN | | • | NVIDIA GTX 1080 Ti | • |
| Dynamic-SLAM [84] | 2019 | RGB | ORB-SLAM2 | SSD | | • | Intel Core i5-7300HQ CPU, NVIDIA GeForce GTX1050Ti GPU | |
| QuadricSLAM [85] | 2019 | RGB | Factor graph | YOLOv3, dual quadrics | | | | • |
| CubeSLAM [86] | 2019 | Mono, stereo RGB | Object SLAM | YOLO, 3D cuboids | | • | | • |
| EM-Fusion [87] | 2019 | RGB-D | Volumetric object SLAM, EM | Mask R-CNN | | • | | • |
| PSPNet-SLAM [61] | 2020 | RGB-D | ORB-SLAM2 | PSPNet | | • | Intel i7 CPU, GTX1070 GPU | |
| DDL-SLAM [63] | 2020 | RGB-D | ORB-SLAM2 | DUNet | | • | Intel i7 CPU, NVIDIA TITAN GPU | |
| SaD-SLAM [57] | 2020 | RGB-D | ORB-SLAM2 | Mask R-CNN | | • | Intel i7-4790 CPU | |
| VDO-SLAM† [24] | 2020 | RGB, Semantic | | Mask R-CNN | • | • | Intel Core i7 2.6 | • |
| SALSA [88] | 2020 | RGB | ORB-SLAM2 | Mask R-CNN | | • | | • |
| DynaSLAM II [89] | 2020 | RGB-D | ORB-SLAM2 | | | • | | |
| NodeSLAM [90] | 2020 | RGB-D | Latent-shape object SLAM | Mask R-CNN, neural object descriptors | | | | |
| RDS-SLAM [23] | 2021 | RGB | ORB-SLAM3 | SegNet and Mask R-CNN | • | • | NVIDIA GeForce RTX 2080Ti GPU | • |
| RDMO-SLAM [51] | 2021 | RGB-D | ORB-SLAM3 | Mask R-CNN | • | • | NVIDIA GeForce RTX 2080Ti | |
| [53] | 2021 | RGB | | Mask R-CNN | | • | | |
| DS-SLAM-HERO [91] | 2021 | RGB-D | DS-SLAM | SegNet | | • | Intel Arrial 10 FPGA, Intel OpenCL SDK v17.1 | • |
| [48] | 2021 | RGB-D | ORB-SLAM2 | SegNet | • | • | NVIDIA Jetson AGX Xavier | |
| OFM-SLAM [62] | 2021 | RGB-D | ORB-SLAM2 | Mask R-CNN | | • | NVIDIA GeForce GTX-1070 | |
| DP-SLAM [92] | 2021 | RGB-D | ORB-SLAM2 | Mask R-CNN | | • | Intel i5-8265U CPU, NVIDIA GeForce MX150 | |
| Kimera [93] | 2021 | Stereo, IMU | Kimera-VIO/PGO, mesh | 2D segmentation to metric–semantic map | | • | | • |
| DSP-SLAM [94] | 2021 | Mono, stereo, stereo + LiDAR | ORB-SLAM2, object BA | Mask R-CNN, DeepSDF shape priors | • | | i9-9900 K, RTX 2080 8 GB; 10 FPS | • |
| SOLO-SLAM [50] | 2022 | RGB-D | ORB-SLAM3 | SOLOv2 | • | • | NVIDIA-3070TI | |
| WF-SLAM [55] | 2022 | RGB-D | ORB-SLAM2 | Mask R-CNN | | • | Intel I7-8700 K processor, GTX1080TI GPU | • |
| RTSDM [49] | 2022 | RGB-D | ORB-SLAM2 | BiSeNetv2 | • | • | NVIDIA Jetson TX2 | |
| Dynamic-VINS [25] | 2022 | RGB-D, IMU | VINS-Mono [95], VINS-RGBD [96] | YOLOv3 | | • | HUAWEI Atlas200 DK, NVIDIA Jetson AGX Xavier | • |
| DynaVINS [97] | 2022 | RGB, IMU | VINS-Mono | Instance masks, geometry | | • | | • |
| AirDOS [98] | 2022 | RGB, IMU | Articulated-object VI-SLAM | Instance masks | | • | | • |
| FCH-SLAM [54] | 2022 | RGB-D | ORB-SLAM2 | FCHarDNet | | | GTX1650 GPU | |
| Blitz-SLAM [69] | 2022 | RGB-D | ORB-SLAM2 | BlitzNet | | • | | |
| YOLO-SLAM [71] | 2022 | RGB-D | ORB-SLAM2 | Darknet19-YOLOv3 | | • | Intel Core i5-4288U CPU | |
| [99] | 2022 | LiDAR | | RangeNet++ | | | | |
| TwistSLAM++ [100] | 2022 | RGB, LiDAR | TwistSLAM [101], S3LAM [102] | 3DSSD [103], Detectron2 [104] | • | • | | |
| [105] | 2022 | RGB-D | GTSAM [106] | YOLOv4 | | | AMD Ryzen 3950x CPU, NVIDIA RTX 2080 T | |
| $D^2$SLAM [64] | 2022 | RGB-D | ORB-SLAM3 | YOLACT++ | | • | Intel XENON CPU, Nvidia GPU RTX2080 SUPER | |
| DGS-SLAM [107] | 2022 | RGB-D | ORB-SLAM3 | YOLACT++ | | • | Nvidia GPU GTX3070 | |
| SO-SLAM [108] | 2022 | RGB-D | ORB-SLAM2, QuadricSLAM [85] | YOLOv3 | | | Intel Core i5-7200U 2.5 GHz CPU | |
| Changing-SLAM [109] | 2022 | RGB-D | ORB-SLAM3 | YOLOv4 | | • | NVIDIA GPU GTX3070 | |
| [110] | 2022 | RGB-D | ORB-SLAM2 | PanopticFCN [44] | | • | NVIDIA RTX 2080Ti GPU | |
| D-VINS [52] | 2023 | RGB, IMU | VINS-Fusion [111] | YOLOv5 | • | | NVIDIA GEFORCE RTX 3050T | |
| 3DS-SLAM [68] | 2023 | RGB | ORB-SLAM2 | HTx build on 3DETR [112] | | • | i9 CPU, 16 GB RAM, RTX 3070 Ti | • |
| YDD-SLAM [113] | 2023 | RGB-D | ORB-SLAM3 | YOLOv5 | | • | AMD Ryzen 7 4800H CPU, NVIDIA RTX 2060 | |
| SG-SLAM [114] | 2023 | RGB-D | ORB-SLAM2 | YOLACT | | | Intel Core i9-10920X, NVIDIA GeForce RTX 3090 | • |
| PVO [115] | 2023 | RGB | DROID-SLAM [116] | PanopticFPN | | • | GeForce RTX 3090 GPU | • |
| SD-SLAM [117] | 2024 | LiDAR | | RangeNet++ | | • | Intel Core i7-8550 U CPU, NVIDIA GeForce MX150 | |
| YOEC-DSLAM [65] | 2024 | RGB | ORB-SLAM2 | YOLOv5s-SEG | | • | Intel Core i5-10400, NVIDIA GeForce RTX 2080Ti | |
| V3D-SLAM [67] | 2024 | RGB-D | ORB-SLAM2 | YOLOv8 | | • | Intel NUC5i3RYH | • |
| SEO-SLAM [66] | 2024 | RGB-D | GTSAM | GroundingDINO [118], SAM [40] | | | | |
| BY-SLAM [70] | 2024 | RGB-D | ORB-SLAM3 | YOLOv8s-FasterNet | | • | NVIDIA 2080Ti, Intel i9-9900X and ARM Cortex-A72 | |
| Panoptic-SLAM [41] | 2024 | RGB | ORB-SLAM3 | PanopticFPN | | • | Intel i7 CPU, NVIDIA RTX3060 GPU | • |
| PSMD-SLAM [42] | 2024 | RGB, LiDAR, IMU | | PanopticFPN | • | • | AMD 5950X CPU, NVIDIA GeForce 3090 GPU | |
| ConsistentPanopticSLAM [43] | 2024 | RGB-D | | Mask2Former [119] | | | 12700 K CPU, Nvidia RTX3090 | • |
| BE-SLAM [120] | 2024 | RGB, LiDAR | ORB-SLAM2, DSP-SLAM [94] | | | • | 2 x NVIDIA GeForce RTX 4090 GPU | • |
| SAM-SLAM [121] | 2025 | RGB-D | ORB-SLAM3 | SAM | • | • | NVIDIA Jetson Nano | |
| YGC-SLAM [122] | 2025 | RGB | ORB-SLAM2 | YOLOv5 | • | • | Nvidia RTX 3060 GPU | |
| TiMoSLAM [123] | 2025 | RGB | ORB-SLAM2 | YOLOv4 | • | | Intel Core i7-10700 K CPU, NVIDIA RTX 3080 GPU | |
| NGD-SLAM [124] | 2025 | RGB-D | ORB-SLAM3 | YOLO | • | • | MD Ryzen 7 4800H CPU, NVIDIA GeForce RTX 2060 GPU | • |
| **Semantic NeRF SLAM** [2.2] | | | | | | | | |
| iLabel [125] | 2021 | RGB-D | iMAP [126] | User | • | | | |
| FR-Fusion [127] | 2022 | RGB-D | iMAP | User, EfficientNet [128], DINO [129] | • | | | |
| NIDS-SLAM [130] | 2023 | RGB-D | ORB-SLAM3 | Mask2Former [119] | • | | | |
| DNS-SLAM [131] | 2023 | RGB-D | | | • | | NVIDIA 2080ti GPU | • |
| vMAP [132] | 2023 | RGB-D | ORB-SLAM3 | | • | | 3.60 GHz i7-11700 K CPU, NVIDIA RTX 3090 GPU | • |
| RO-MAP [133] | 2023 | RGB, Semantic | ORB-SLAM2 | YOLOv8 | | | Intel Xeon 6154 CPU, NVIDIA RTX 4090 GPU | • |
| SNI-SLAM [27] | 2024 | RGB-D | | Dinov2 [32] | | | NVIDIA RTX 4090 GPU | • |
| NIS-SLAM [134] | 2024 | RGB-D | | Mask2Former | • | | NVIDIA RTX 4090 | |
| DN-SLAM [135] | 2023 | RGB-D | ORB-SLAM3 | SAM | • | • | Intel i9 CPU, RTX4090 GPU, 32 GB of RAM | |
| Dynamon [136] | 2023 | RGB | DROID-SLAM | | | • | NVIDIA RTX 4090, 24 GB of VRAM | |
| DDN-SLAM [137] | 2024 | RGB-D, Semantic | ORB-SLAM3 | YOLOv9 | • | • | NVIDIA RTX 3090ti GPU, Intel Core i7-12700 K CPU 3.60 GHz | |
| NID-SLAM [138] | 2024 | RGB-D, Semantic | | YOLO | • | • | 3.20 GHz Intel Core i9-12900KF CPU, NVIDIA RTX 3090 GPU | |
| DVN-SLAM [139] | 2024 | RGB-D | | | • | • | NVIDIA A100 GPU | |
| RoDyn-SLAM [140] | 2024 | RGB-D | Dynamic neural radiance field | Semantic and optical-flow masks | • | • | | • |
| $SLC^2$-SLAM [141] | 2025 | RGB-D | Co-SLAM | SemMet | • | | AMD Ryzen 9 5950X CPU, NVIDIA GeForce RTX 4090 GPU | |
| FND-SLAM [142] | 2025 | RGB-D | ESLAM [143] | YOLOv8 | • | • | | |
| DMN-SLAM [144] | 2025 | RGB-D | ORBSLAM2 | LSSMask | • | • | | |
| SNI-SLAM++ [145] | 2026 | RGB-D | Tightly coupled neural implicit SLAM | Hierarchical semantic features | • | | | |
| NeuSE [146] | 2026 | RGB-D | Long-term object-level SLAM | SE(3)-equivariant object embeddings | | | | |
| **Semantic Gaussian Splatting SLAM** [2.3] | | | | | | | | |
| SGS-SLAM† [28] | 2024 | RGB-D, Semantic | | | • | • | NVIDIA A100-40 GB GPU | • |
| SemGauss-SLAM [147] | 2024 | RGB-D | | Dinov2 [32] | | | NVIDIA RTX 4090 GPU | |
| NEDS-SLAM [148] | 2024 | RGB-D | | Depth Anything [149] | • | | NVIDIA RTX 4090 GPU | |
| GS3LAM† [26] | 2024 | RGB-D, Semantic | | | • | | AMD EPYC 7302 16-CORE Processor, NVIDIA RTX 3090 GPU | • |
| Hi-SLAM [150] | 2024 | RGB-D, Semantic | | | | | Nvidia L40S GP | |
| Pano-SLAM [151] | 2024 | RGB-D | SplaTAM [152] | SEEM [45] | | | RTX 4090 GPU | |
| DG-SLAM (Xu et al.) [153] | 2024 | RGB-D | DROID-VO + 3DGS | OneFormer, depth warp | | • | RTX 3090 Ti | • |
| MG-SLAM [154] | 2024 | RGB | DPVO + 3DGS | CLIP loop descriptors | • | | i7-12700 K, RTX 4090 | |
| GS4 [155] | 2025 | RGB-D | ODIN [156] | ODIN [156] | • | | Nvidia RTX TITAN | |
| Hier-SLAM++ [157] | 2025 | RGB-D | | GPT-4 Turbo | | | RTX 4090, L40S | |
| RGBDS-SLAM [158] | 2025 | RGB-D, Semantic | ORB-SLAM3 | | • | | NVIDIA RTX 3060 GPU, AMD Ryzen 7 5800H CPU | • |
| OpenGS-SLAM [159] | 2025 | RGB-D | | RAM [160], SAM | • | | | • |
| STAMICS [161] | 2025 | RGB-D | | | • | | | |
| SDD-SLAM [162] | 2025 | RGB-D | | GroundingDINO, SAM-Track [163] | • | • | NVIDIA RTX 3080 GPU | |
| GSFF-SLAM [164] | 2025 | RGB-D | | | • | | NVIDIA RTX 4090 GPU | |
| SGF-SLAM [165] | 2025 | RGB | Photo-SLAM | STDC-seg [166] | | • | Intel i5-13,600 KF CPU, 32 GB RAM, NVIDIA RTX 3090 Ti | |
| DGS-SLAM [167] | 2025 | RGB-D | | | • | • | | • |
| Gassidy [168] | 2025 | RGB-D | | YOLO | • | • | Intel i9-12900 K CPU, NVIDIA RTX3080 GPU | |
| GARAD-SLAM [169] | 2025 | RGB-D | | | • | • | Intel Core i9-14900 K CPU, RTX 4080 Ti GPU | |
| DyGS-SLAM [170] | 2025 | RGB-D | ORB-SLAM3 | EfficientSAM, YOLO-World | • | • | | |
| DynaGSLAM [171] | 2025 | RGB-D | DynoSAM [172] | | | • | NVIDIA 3090Ti | |
| WildGS-SLAM [173] | 2025 | RGB | DROID-SLAM | DINOv2 | • | • | | • |
| Embracing Dynamics [174] | 2025 | RGB-D | | | • | • | Intel Core i7-13700HX CPU, NVIDIA GeForce RTX 4080 GPU | |
| Dy3DGS-SLAM [175] | 2025 | RGB | | lightweight U-Net | • | • | RTX 3090 Ti GPU | |
| CAD-SLAM [176] | 2025 | RGB-D | | | • | • | Intel Platinum 8362 CPU, NVIDIA A100 GPU | |
| UP-SLAM [177] | 2025 | RGB-D | | DINOv2 | • | • | Intel i7-12700KF, NVIDIA RTX 4060ti 16G GPU | • |
| OVO (Gaussian-SLAM) [178] | 2025 | RGB-D | ORB-SLAM, Gaussian-SLAM | SAM 2.1-L, SigLIP | • | | i7-11700 K, RTX 3090, 64 GB | • |
| Online Language Splatting [179] | 2025 | RGB-D | Online 3DGS-SLAM | CLIP, online autoencoder | | | | |
| DQO-MAP [180] | 2026 | RGB-D | ORB-SLAM2, quadrics + 3DGS | YOLOv10 object detections | • | | | • |
| Open-Set Semantic GS-SLAM [181] | 2026 | RGB-D | SplaTAM, LoopSplat extension | SAM, CLIP; expandable feature pool | • | | RTX 3090, 19.5 GB reported | |
| TGS-SLAM [182] | 2026 | RGB-D, Semantic | 3DGS + TriDS | Precomputed semantic labels | | | | • |
| ReSemGS-SLAM [183] | 2026 | RGB-D | Semantic-consistency 3DGS SLAM | Semantic perception model | | | | |

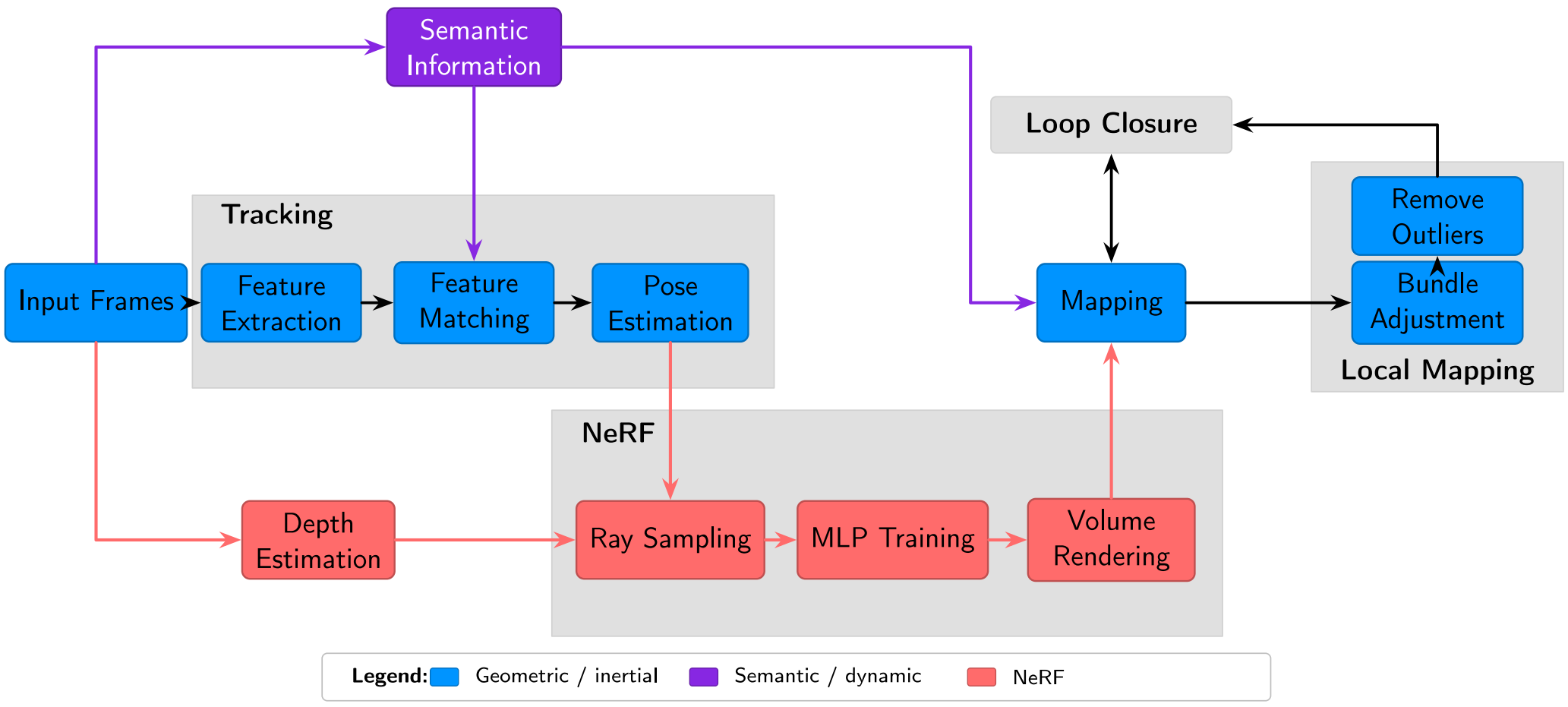

**Fig. 6.** Overview of NeRF-enhanced SLAM architecture. The system integrates a NeRF module (in coral) with a geometric SLAM pipeline (in blue). Starting from input frames, pose estimates are computed using standard tracking modules. The NeRF component receives these pose estimates and uses depth and RGB data to train a NeRF, enabling high-quality dense mapping and novel view synthesis. This integration leverages the strengths of both geometric SLAM for robust pose estimation and NeRF for volumetric scene representation. Semantic information (in purple) can be incorporated at different stages (either as input data or through a parallel thread) in the architecture, depending on the specific approach.

FR-Fusion [127] extends iMAP's efficient neural field architecture by incorporating semantic capabilities through latent feature fusion. The system processes RGB input using pre-trained networks such as EfficientNet [128] or DINO [129] to extract high-dimensional features (up to 1536D). These features are combined using a latent volumetric rendering technique that minimizes the computational overhead associated with high-dimensional data. To maintain real-time performance, FR-Fusion employs guided keyframe and pixel sampling, enabling efficient mapping and semantic understanding. The fusion process ensures consistent neural field reconstruction while supporting open-set semantic segmentation.

RO-MAP [133] leverages a dual-component architecture that combines SLAM with neural implicit representations. The first component implements a lightweight object SLAM atop ORB-SLAM2, utilizing instance segmentation to detect objects and estimate their pose through multi-view geometry. A robust data association algorithm ensures correct mapping of observations across frames. The second component handles reconstruction by maintaining separate NeRF models for each detected object. Each object instance receives its own neural radiance field, trained incrementally as new observations become available. The training process employs a sampling strategy that selects keyframes based on significant viewpoint changes, while a custom loss function combines photometric consistency with geometric constraints. To handle background interference, the system implements ray classification based on instance masks and applies specific optimization objectives for object surfaces versus empty space. The parallel training architecture uses a thread pool approach where each worker maintains a dedicated CUDA stream for asynchronous optimization of object models.

While these initial approaches demonstrated the potential of semantic NeRF-SLAM, the computational complexity of feature fusion led to the exploration of unified representations. iLabel [125] introduces a unified neural representation that combines geometry, appearance, and semantics in a single MLP network. The system operates by continuously optimizing this representation as users provide sparse click annotations in keyframes. The network learns to propagate these annotations through geometric and appearance similarities encoded in its weights. The system implements both flat and hierarchical semantic modes, with the latter enabling tree-structured scene understanding. The optimization process uses volumetric rendering with carefully designed activation functions and loss terms to ensure smooth label propagation while maintaining geometric accuracy.

Following this trend of unified representation, but addressing its limitations in handling inconsistent inputs, NIDS-SLAM combines traditional ORB-SLAM3 tracking with implicit neural mapping. The system's key innovation is its semantic color encoding scheme, which transforms semantic labels into color values for efficient fusion. The mapping process optimizes a dynamic set of keyframes, with semantic information encoded in feature planes. The system employs driven by data feature extraction and uncertainty-based keyframe selection to maintain computational efficiency. The decoder architecture enables separate optimization of geometry and semantics while maintaining their correlation, allowing robust semantic mapping even with inconsistent input segmentation.

The challenge of maintaining real-time performance while ensuring semantic consistency led to the development of hybrid approaches. DNS SLAM [131] introduces a neural RGB-D semantic SLAM system that simultaneously performs camera tracking, dense reconstruction, and semantic understanding. The system's key innovation lies in its hybrid representation combining multi-resolution hash-grid based features with positional encoding, enabling both detailed geometry reconstruction and consistent semantic modeling. The approach handles noisy 2D semantic segmentation input through a multi-view semantic fusion strategy that aggregates information across frames to achieve 3D consistent semantics. DNS SLAM is especially notable for preserving high-quality reconstruction and semantic consistency, achieving state-of-the-art tracking accuracy with more than a 10% improvement over previous methods. However, its reliance on RGB-D input and pre-trained 2D segmentation models may limit its applicability in real-world scenarios.

While previous methods focused on scene-level semantics, vMAP [132] takes a fundamentally different direction by addressing object-level understanding. This system represents a significant advancement in object-level dense SLAM by introducing separate tiny MLP neural field models for each object in the scene, enabling efficient and watertight object modeling without requiring 3D priors. The system's key contribution is its vectorized training approach that allows simultaneous optimization of up to 50 individual object models on a single GPU. This object-centric approach achieves superior scene-level and object-level reconstruction quality compared to previous neural field SLAM systems, while maintaining efficient memory usage at just 40 KB per object. The system demonstrates robust performance in real-time operation, although it currently relies on external tracking systems for pose estimation, which could be a limitation for fully autonomous embedded applications.

Building upon the success of hybrid representations, but addressing their computational limitations, NIS-SLAM [134] presents an efficient implicit neural semantic RGB-D SLAM system that integrates 2D semantic priors to achieve consistent 3D scene understanding. The system employs a novel hybrid representation combining tetrahedron-based features with positional encoding, alongside a multi-view semantic fusion strategy to handle inconsistent segmentation results. A notable contribution is its confidence-based pixel sampling and progressive optimization weight function for robust camera tracking. The system achieves competitive performance in both synthetic and real-world environments, although its computational requirements may pose challenges for limited resources embedded systems. The approach demonstrates particular strength in maintaining semantic consistency across multiple views while performing real-time tracking and mapping.

Recent NeRF-based SLAM systems, developed since 2023, are now capable of handling dynamic scenes, effectively managing moving objects while maintaining accurate scene reconstruction. These systems employ a variety of strategies, such as dynamic object segmentation, real-time scene adaptation, and motion prediction, to ensure robust performance in environments with frequent changes. DN-SLAM [135] combines ORB features with NeRF mapping using a two-stage segmentation process that integrates optical flow and SAM for dynamic region detection. DynaMoN [136] implements dynamic content masking with HexPlane's 4D space–time grid structure, focusing ray sampling on static regions during optimization. DDN-SLAM uses YOLOv9 detection coupled with Gaussian Mixture Models to segment dynamic features, incorporating a background restoration strategy based on optical flow and depth values. NID-SLAM [138] focuses on depth-guided semantic mask enhancement and selective keyframe strategies that minimize dynamic object interference. DVN-SLAM [139] introduces a fusion approach between discrete feature planes and continuous neural radiance fields, using attention mechanisms to maintain the structure of the scene. These systems share core architectural elements: they all implement dynamic object detection through semantic or geometric analysis, separate tracking from mapping processes, and provide mechanisms to reconstruct occluded backgrounds. Their main differences lie in scene representation choices and their specific approaches to dynamic content handling, ranging from optical flow analysis to probabilistic models.

The development of NeRF-based Semantic SLAM systems highlights substantial progress in neural implicit representations, enabling the simultaneous mapping of geometric structures and semantic information. The field has evolved from basic feature integration methods to sophisticated unified and hybrid approaches, each offering unique trade-offs between computational efficiency, semantic accuracy, and real-time performance. However, the use of NeRF in SLAM is not without its challenges. The main computational bottleneck is not only the size of the MLP, but the way it is repeatedly evaluated and optimized. Tracking and mapping require volumetric rendering along many camera rays; each ray is sampled at multiple depth locations, and each sample triggers neural field queries for density, color, and often semantic features. During online mapping, these queries are followed by backpropagation through photometric, geometric, and semantic losses, while keyframe replay is needed to avoid forgetting and maintain global consistency. Semantic NeRF-SLAM therefore combines MLP or decoder training, ray sampling, differentiable rendering, pose refinement, and semantic feature fusion in the online loop. This creates high latency, large activation memory during training, and strong pressure on memory bandwidth. On embedded GPUs, where power and memory bandwidth are limited, these operations make NeRF-based Semantic SLAM difficult to deploy in real time, even when compact encodings or keyframe sampling are used. With ongoing research into optimization techniques and more efficient neural network models, NeRF-integrated semantic SLAM systems have the potential to offer substantial improvements over Geometric SLAM, but their current online training and rendering cost remains a major barrier for embedded deployment.

### 2.3. Semantic Gaussian splatting SLAM

The limitations of NeRF have spurred significant research into alternative representations. Introduced by Kerbl et al. [185] in 2023, 3D Gaussian Splatting provides an explicit radiance field approach tailored for efficient and high-quality rendering of 3D scenes. In contrast to traditional volumetric representations, such as voxel grids, 3DGS employs differentiable 3D Gaussian primitives, which offer a continuous, memory-efficient, and flexible way to model complex spatial and radiance distributions. Using the compact and smooth properties of Gaussians, this method achieves real-time rendering capabilities on modern GPUs, while also offering enhanced scalability and adaptability to dynamic scenes. These advancements have led to substantial progress in Gaussian Splatting-based SLAM systems, which offer superior performance and reconstruction quality compared to traditional neural implicit approaches. Gaussian Splatting SLAM operates through four key components: initialization, tracking, mapping, and optimization, offering a tightly integrated approach to both geometry and semantics (Fig. 7). Unlike traditional SLAM systems, where semantic information is typically added as an independent module, this system seamlessly embeds semantic information directly into the primary pipeline. The process begins with the initialization of sparse Gaussian representations derived from the first few camera frames. These Gaussians serve as the foundation for the scene and are progressively enriched with semantic labels as new frames are introduced. In the tracking phase, the camera pose is estimated relative to the Gaussian map. This involves aligning observed features with the 3D Gaussian representation using traditional SLAM techniques. Pose estimation combines two key approaches: geometric alignment, which minimizes reprojection errors of Gaussian splats, and photometric alignment, which refines the pose by reducing intensity and color differences between observed frames and the Gaussian map. During mapping, the system refines each Gaussian parameter, such as position, size, and orientation, while simultaneously updating semantic attributes. This creates a unified representation that combines geometric precision with semantic awareness. New Gaussians are added as needed, ensuring that the map evolves dynamically. The optimization phase jointly adjusts the camera poses and Gaussian parameters, ensuring global consistency between the map and observed frames.

The groundwork for semantic integration in 3DGS was laid by SGS-SLAM [28], which introduced the integration of semantics with 3DGS through a multi-channel optimization approach. The system jointly optimizes appearance, geometry, and semantic features through differentiable rendering, achieving state-of-the-art semantic accuracy. A key innovation is its depth-adaptive scale regularization scheme that prevents over-smoothing while maintaining semantic consistency. The system leverages semantic information for keyframe selection and mapping optimization (i.e. loop closure detection), allowing downstream tasks such as scene editing. Enhancing this foundation through a more sophisticated feature representation strategy, SemGauss-SLAM [147] directly embeds high-dimensional semantic features extracted through Dinov2 [32] into the Gaussian representation. The system propagates 2D semantic features to initialize 3D Gaussians and employs feature-level loss alongside traditional RGB/depth losses for optimization. A distinctive contribution is its semantic-informed bundle adjustment that leverages feature consistency across co-visible frames to reduce drift and improve reconstruction quality. This approach enhances the synthesis of novel semantic views and improves memory efficiency through feature compression, although the feature extraction process remains computationally demanding. Although the first approaches emphasized accuracy over efficiency, more recent methods have changed focus to meet the constraints of deployment. NEDS-SLAM [148] proposes an architecture based on its Spatially Consistent Feature Fusion (SCFF) model and lightweight encoder–decoder design. The system combines semantic features from pre-trained models with appearance features from Depth Anything [149] to reduce inconsistency in semantic estimates. Its Virtual Camera View Pruning (VCVP) method efficiently identifies and removes outlier Gaussians by using multiple viewpoints. Notably, NEDS-SLAM achieves competitive accuracy with an interesting reduced memory footprint and demonstrates viable performance on mid-range GPUs.

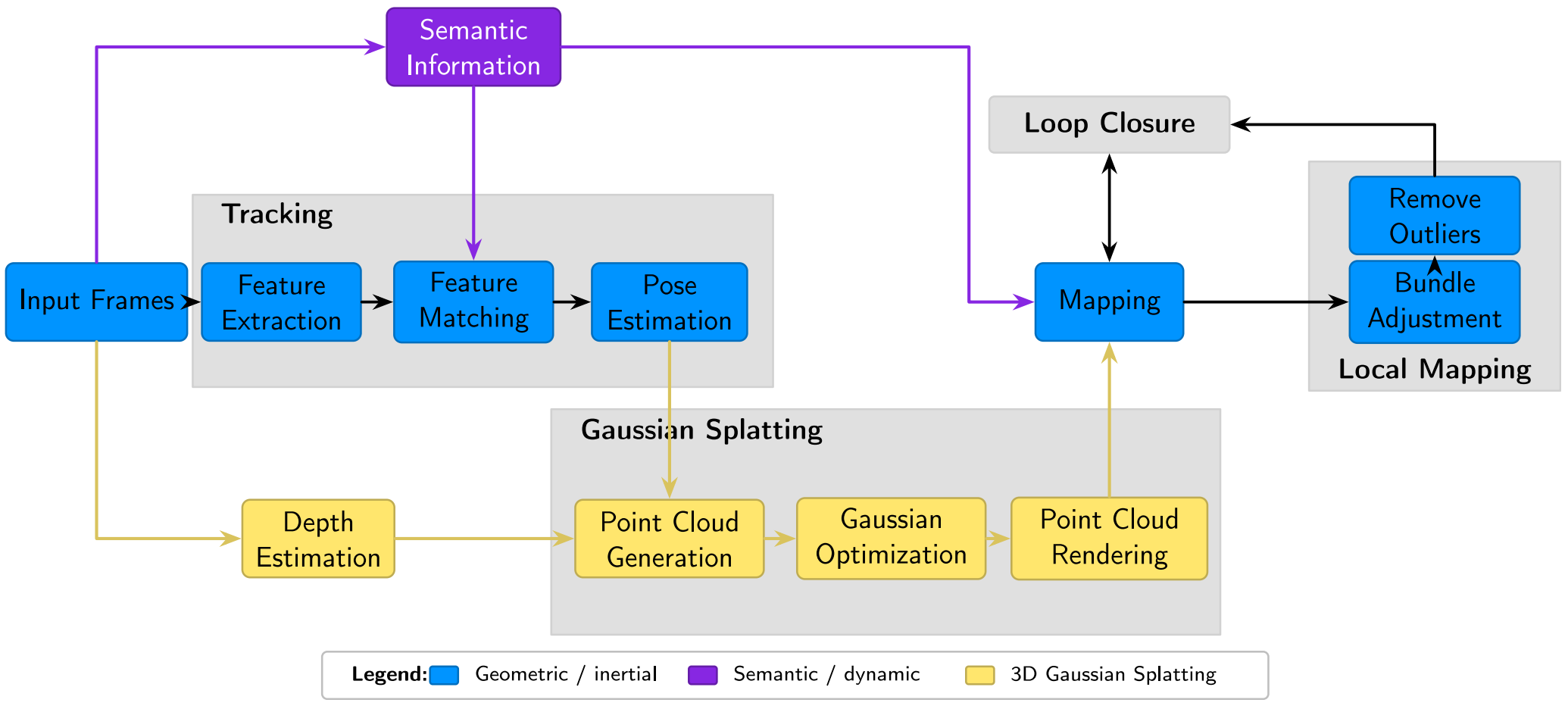

**Fig. 7.** Overview of Gaussian Splatting-enhanced SLAM architecture. This system enhances traditional SLAM (in blue) with a Gaussian Splatting module (in yellow). After estimating poses using the SLAM pipeline, the GS component generates a point cloud from RGB-D inputs, initializes 3D Gaussians, and optimizes their parameters (position, scale, and color). These optimized Gaussians are used for dense mapping and rapid image rendering, providing an explicit geometric representation while benefiting from the robust pose estimation of the SLAM system. Like in NeRF-based SLAM, semantic information (in purple) can be integrated at various points in the architecture depending on the approach.

In parallel to efficiency-focused approaches, some methods have explored novel semantic representation strategies. GS3LAM [26] introduces the concept of Semantic Gaussian Fields, explicitly modeling semantic information within the Gaussian framework. The system's Random Sampling-Based Keyframe Mapping strategy effectively addresses the forgetting phenomenon in incremental mapping while maintaining semantic consistency. Although it provides strong semantic reconstruction and tracking accuracy, its high memory demands and computational complexity make it better suited for high-performance platforms than for embedded systems. However, its field-based representation provides valuable insight for future optimization strategies in semantic 3DGS-SLAM systems.

A significant advancement in semantic representation comes from Hi-SLAM [150], which introduces a hierarchical approach to semantic organization. Its architecture revolves around two main components: (1) an LLM-assisted semantic tree generation that organizes semantic classes into a hierarchical structure, significantly reducing memory footprint, and (2) a hierarchical loss function that combines inter-level and cross-level optimization to maintain semantic consistency at different granularity scales. The pipeline alternates between tracking steps, where camera poses are estimated using differentiable rendering, and mapping steps, where the 3D Gaussian map is optimized by incorporating both geometric and semantic information. Although 3DGS avoids the dense MLP queries required by NeRF and can render efficiently on desktop GPUs, Semantic GS-SLAM introduces other bottlenecks. Each mapping step optimizes the position, scale, rotation, opacity, appearance, and semantic attributes of a growing set of Gaussians. Densification, pruning, visibility sorting, tile-based rasterization, and optimizer states increase both computation and memory traffic. The semantic extension further enlarges each primitive because semantic logits or feature vectors must be stored, rendered, and optimized together with geometric and photometric parameters. Consequently, 3DGS shifts the bottleneck from implicit-field evaluation to large-scale differentiable rasterization and Gaussian-parameter optimization, which remains challenging on embedded GPUs.

## 3. Embedded semantic SLAM

The deployment of Semantic SLAM systems on embedded platforms presents unique challenges and opportunities in robotics applications. Embedded systems, characterized by their specialized computing architecture and resource constraints, require careful consideration of computational efficiency, memory usage, and power consumption. The emergence of heterogeneous computing platforms that incorporate GPUs and neural processing units (NPUs) has enabled the deployment of deep learning models in real-time applications. Contemporary platforms, including the NVIDIA Jetson family (such as the AGX Orin), Qualcomm Robotics RB5, Google Coral Edge TPU, Intel Movidius Myriad, AMD Kria SoC, or Texas Instruments TDA4VM, offer the computational infrastructure required for edge AI applications while prioritizing power efficiency.

This has driven extensive work into adapting SLAM algorithms for low-power embedded systems, reengineering them to be compatible with these architectures. Research efforts have focused on optimizing various components of the SLAM pipeline, from VO to mapping and semantic understanding. Some researchers [186,187] have concentrated on VO, prioritizing it over SLAM but with a trade-off in reduced accuracy. Several approaches have shown success in the implementation of based on keyframes SLAM systems on platforms with limited resources [48,84,121,188,189]. Visual–inertial odometry (VIO) systems have particularly benefited from hardware acceleration strategies, notably through Field Programmable Gate Arrays (FPGA) implementations [91,190–192]. Specialized hardware architectures have emerged to address the specific requirements of visual SLAM systems. The Navion processor [193] allows to implement a complete based on keyframes VIO pipeline on a single chip, while the CNN-SLAM processor [194] accelerates CNN-based feature extraction and matching operations.

A notable advancement in hardware–software co-design is demonstrated in [195], where on-sensor Optical Flow estimation is integrated with the VINS-Mono pipeline. This approach utilizes an ASIC-based accelerator for feature tracking while executing the remaining VIO components on a Raspberry Pi Compute Module 4. In the domain of LiDAR-based semantic SLAM, recent work [196] has introduced an energy-efficient System-on-Chip (SoC) designed to address the computational challenges of point neural networks and LiDAR odometry. The architecture features a multi-granularity parallel multi-core design connected through a 2D-mesh Network-on-Chip (NoC). Each type of core is specialized for specific Semantic LiDAR-SLAM tasks, and most cores incorporate SIMD processing elements for parallel computation. The nonlinear optimization core employs a reconfigurable architecture to optimize utilization through instruction pipelining and variable SIMD modes. In [197], a real-time and fully integrated Semantic LiDAR SLAM processor (LSPU) is presented with the semantic LiDAR-PNN-SLAM (LP-SLAM) system, which provides 3D segmentation, localization, and mapping simultaneously based on the point neural network.

Despite these advances, the implementation of semantic SLAM on embedded platforms still faces significant challenges, particularly in balancing algorithmic sophistication with resource constraints. The integration of semantic understanding capabilities substantially increases computational and memory requirements, making efficient implementation particularly challenging on limited resources systems. To the best of our knowledge, the current state-of-the-art covering object- or semantic-aware SLAM on embedded systems is characterized by four main approaches. These systems are primarily distinguished by how they manage semantic information and implement computational optimization strategies. RTSDM (Real-Time Semantic Dense Mapping) [49] demonstrates an efficient approach for UAV applications, based on the ORB-SLAM2 framework. The system introduces a direct feature matching method that reduces computational overhead compared to traditional keypoint-based approaches. Semantic understanding is achieved through a lightweight deep learning model, particularly BiSeNetV2, with semantic segmentation selectively applied to keyframes. The system employs OctoMap for efficient 3D mapping, using its hierarchical structure for probabilistic occupancy and semantic information storage. Although real-time performance is achieved on the NVIDIA Jetson TX2 platform, the system's inability to handle dynamic objects underscores the persistent challenges in embedded Semantic SLAM. In this context, Semantic SLAM approaches for embedded systems can be broadly classified according to their design and optimization strategies: (i) lightweight geometric SLAM methods with selective semantic integration (e.g., keyframe-based approaches), (ii) systems leveraging efficient deep learning models for real-time semantic processing, (iii) hardware-accelerated implementations using FPGA, ASIC, or heterogeneous architectures, and (iv) hybrid approaches combining algorithmic and hardware-level optimizations. These approaches highlight several key challenges specific to embedded Semantic SLAM, including limited computational resources, constrained memory, high energy consumption, and strict real-time requirements. In addition, the integration of semantic models introduces significant overhead, making the trade-off between semantic accuracy and computational efficiency a central issue for embedded deployment.

A significant advancement in handling dynamic environments is presented in [48], which introduces one of the first real-time semantic RGB-D SLAM systems for low-power platforms. The system employs an efficient dynamic object detection module based on reprojection errors and selective semantic segmentation for keyframes only. The innovative geometry module enables the detection of unknown moving objects through depth image clustering and dynamic area identification. This approach achieves a balance between accuracy and computational efficiency, demonstrating real-time performance on the NVIDIA Jetson AGX Xavier. Dynamic-VINS [25] presents another approach to real-time operation in dynamic environments (Fig. 8), focusing on RGB-D inertial odometry. The system combines object detection through YOLOv3 with depth information for dynamic feature recognition, while utilizing IMU data for motion prediction and feature tracking consistency. Although limited to odometry, the system's efficient handling of dynamic environments provides valuable insights for embedded SLAM implementation.

Optimization of Semantic-aware Geometric SLAM through hardware acceleration is demonstrated in [91], where DS-SLAM is enhanced using OpenCL (Open Computing Language)-based FPGA acceleration. The system addresses the computational intensity of semantic segmentation by implementing SegNet on an FPGA accelerator, enabling efficient execution on the Intel Arria 10 FPGA and the Intel OpenCL SDK v17.1. This hardware–software co-design approach exemplifies the potential for optimizing semantic SLAM systems for embedded deployment while maintaining algorithmic capabilities. A more integrated co-design strategy has been investigated for Semantic LiDAR SLAM. The low-power processor proposed in [196] jointly supports point neural network (PNN) inference, LiDAR odometry, and mapping. It combines a multi-granularity parallel multicore architecture, dedicated $k$-nearest-neighbor cores based on spherical-bin searching, a global point-level scheduler for PNN workload balancing, and a hierarchical reconfigurable aggregation unit. The resulting SoC achieves more than 20 FPS while consuming 164.9 mW and 8.245 mJ per frame. The LSPU processor [197] further extends this approach by integrating semantic segmentation, localization, mapping, and nonlinear optimization within the same hardware platform. Dedicated processing units are used for point-cloud neighborhood search, keypoint extraction, neural-network inference, and pose optimization, resulting in a reported processing time of 20.7 ms. These works illustrate two complementary acceleration schemes: module-level offloading of semantic inference in visual Semantic SLAM and pipeline-level integration of perception, localization, mapping, and optimization in Semantic LiDAR SLAM. However, comparable end-to-end joint hardware and algorithm design remains largely unexplored for visual Semantic SLAM.

## 4. Experiments and analysis

For this comparison study, a variety of Semantic SLAM systems has been chosen, including both Geometric SLAM enhanced with semantic information and recent approaches based on Gaussian Splatting. Among the Semantic Geometric techniques, we opted for RDS-SLAM, which is built on the widely used ORB-SLAM2 framework, and integrated with Mask-RCNN and SegNet models, reflecting different trade-offs in semantic processing complexity. We also included VDO-SLAM and Dynamic-VINS, both designed to handle dynamic environments. In the evaluated VDO-SLAM configuration, the semantic masks are generated through offline Mask R-CNN preprocessing and subsequently supplied to the SLAM pipeline. Its measured runtime therefore covers the use of these masks during tracking and mapping, but excludes their generation. Dynamic-VINS, in contrast, was specifically designed for resource-constrained platforms and performs its semantic processing online. In addition, we included two advanced Gaussian Splatting-based Semantic SLAM systems, GS3LAM and SGS-SLAM. Both consume semantic masks or labels prepared before online SLAM execution. Their runtime measurements therefore include tracking, mapping, semantic fusion, and Gaussian-map optimization, but exclude semantic inference. According to the state of the art, GS3LAM achieves top-tier semantic performance, delivering the highest Mean Intersection over Union (mIoU) among the systems studied. On the other hand, SGS-SLAM extends the capabilities of Gaussian Splatting by introducing robust dynamic object handling. This selection enables a comprehensive evaluation of various strategies to identify the most suitable solutions for limited resources systems, with a focus on balancing performance and computational efficiency.

We test Semantic SLAM systems using their default settings, without any optimization of neural network components. Thus, the performance metrics were obtained using PyTorch inference, without leveraging TensorRT. This choice reflects the out-of-the-box portability of current implementations on embedded hardware; additional platform-specific optimization could improve absolute runtime, but would not remove the architectural cost of semantic inference, dense mapping, or large scene representations. All systems are evaluated using several well-known metrics, including pose accuracy and segmentation accuracy, while also examining runtime (FPS), power consumption (W), and GPU memory utilization to assess their efficiency and practicality in real-world applications. The Absolute Trajectory Error (ATE) is used to evaluate localization accuracy by measuring the average translational difference between estimated camera poses and ground truth trajectories. This metric effectively reflects the system's ability to maintain consistent localization and highlights potential drift issues in various environments. For semantic understanding, the mIoU quantifies segmentation performance by computing the overlap between predicted and ground-truth masks for each object class.

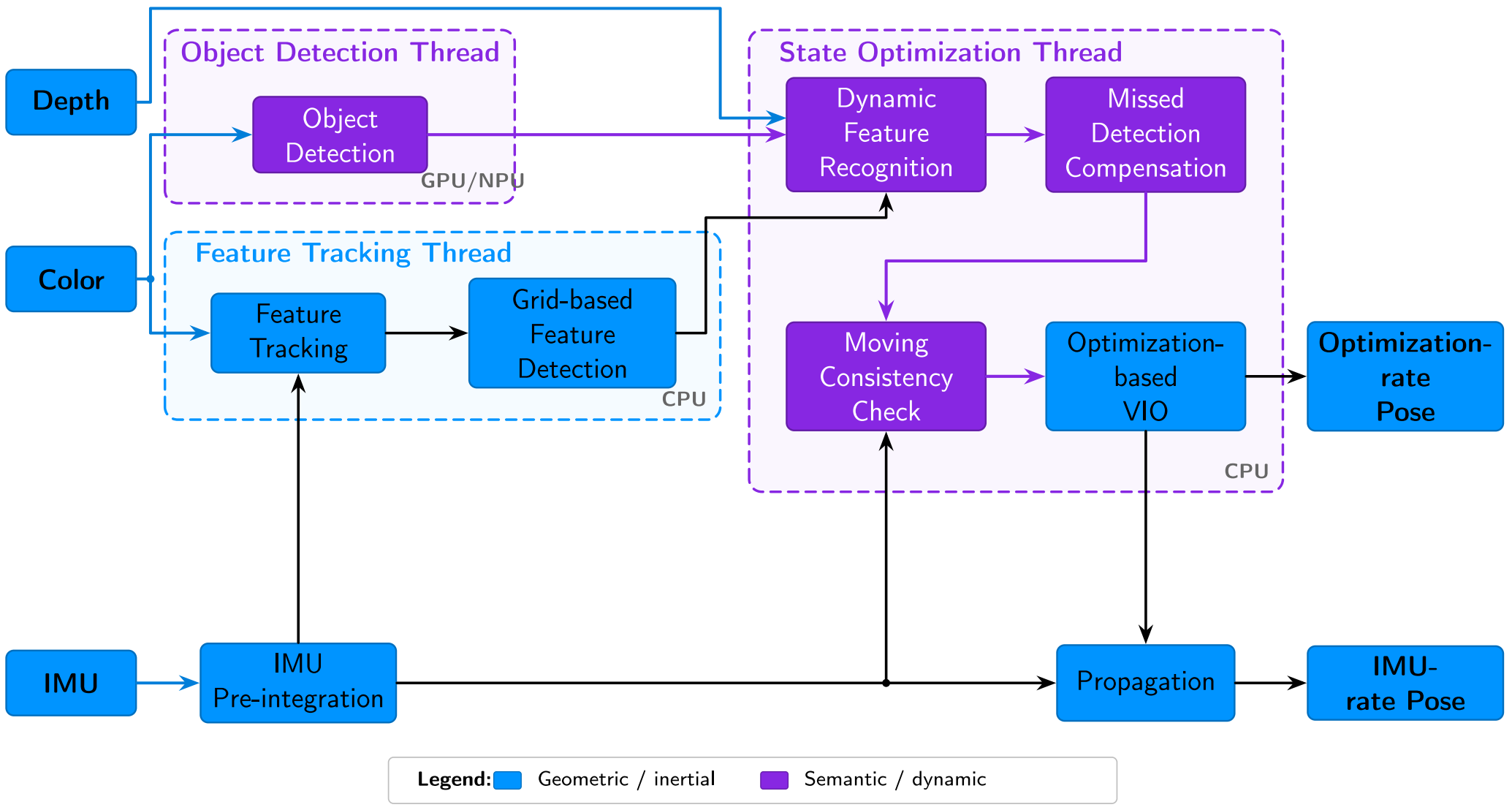

**Fig. 8.** Overview of the Dynamic-VINS architecture [25]. Three threads run in parallel: object detection on the GPU/NPU, feature tracking on the CPU, and state optimization on the CPU. Color frames feed detection and tracking, while depth, detections, and grid-based features are fused for dynamic-feature recognition. IMU pre-integration assists feature tracking and the moving-consistency check. Optimization-based VIO produces an optimization-rate pose, whereas inertial propagation provides an IMU-rate pose. Geometric and inertial modules are shown in blue, and semantic or dynamic-processing modules in purple.

### 4.1. Experimentation setup

The experiments were conducted on the NVIDIA Jetson AGX Orin 64 GB development kit. This embedded platform combines a 2048-core NVIDIA Ampere architecture GPU featuring 64 Tensor Cores with a 12-core ARM Cortex-A78AE CPU running at 2.2 GHz. The 64 GB LPDDR5 memory and AI performance capability of 275 TOPS (INT8) provide the necessary resources to evaluate sophisticated SLAM algorithms while maintaining the power and size constraints typical of robotics applications. To ensure experimental reproducibility, we containerized all SLAM implementations using Docker. The ARM architecture requires many libraries and software components to be recompiled. Algorithms and tools are often optimized for NVIDIA CUDA-enabled GPUs on x86 platforms, creating compatibility and performance barriers when ported to ARM. This setup therefore allows us to assess not only algorithmic accuracy, but also deployability constraints that are invisible in desktop-GPU evaluations, including portability effort, peak memory footprint, sustained power draw, and whether semantic information is computed online or loaded from preprocessing. This provides consistent testing environments and enables reliable performance comparisons between the different systems under evaluation.

For a fair interpretation of the results, we distinguish complete online Semantic SLAM pipelines from SLAM pipelines supplied with semantic inputs precomputed offline. VDO-SLAM, GS3LAM, and SGS-SLAM belong to the latter category. For these methods, the reported runtime, memory consumption, power consumption, and FPS include the loading and downstream processing of semantic information, but exclude semantic inference. Their measurements therefore characterize the executed SLAM loop rather than a complete end-to-end Semantic SLAM pipeline.

### 4.2. Dataset

Evaluating Semantic SLAM requires datasets that allow the assessment of both geometric accuracy and semantic understanding. For these experiments, we use two widely used datasets: TUM RGB-D [198] and Replica [199]. Our evaluation strategy leverages these datasets' complementary strengths: while TUM RGB-D tests robustness against dynamic interference through its carefully curated sequences with moving objects, Replica enables assessment of semantic-aware approaches, particularly for novel rendering techniques like NeRF and Gaussian Splatting.

TUM RGB-D provides thirty-nine RGB-D sequences captured by Microsoft Kinect in indoor environments, with ground-truth trajectories from a motion capture system. What makes TUM RGB-D particularly valuable is its inclusion of challenging dynamic scenarios: people walking through the scene, objects being moved, and furniture being rearranged. These dynamic elements, combined with the varying motions of the camera and lighting conditions, create realistic test cases that mirror the challenges faced by real-world robotic applications. Sequences like `fr3/walking_xyz` and `fr3/sitting_xyz` specifically focus on human motion, while `fr3/walking_static` includes both moving people and static scenes, allowing systematic evaluation of a SLAM system's ability to handle dynamic objects. To ensure a robust and fair performance evaluation, the results will be reported as averages across these sequences, which are diverse enough to cover a range of scenarios, from static to highly dynamic environments. However, the dataset lacks semantic annotations, limiting its use to purely geometric SLAM evaluation.

Replica provides high-fidelity reconstructions of 18 indoor scenes with dense semantic annotations in 88 object categories. Despite being synthetic, its comprehensive semantic labels and accurate reconstructions make it particularly suitable for evaluating novel rendering approaches like NeRF and 3D Gaussian Splatting that can integrate semantic information into their scene representations. The semantic annotations of the dataset enable a quantitative evaluation of the semantic understanding capabilities along with the quality of the geometric reconstruction.

**Table 3**
Trajectory accuracy on the TUM RGB-D dataset ordered by increasing ATE RMSE. Values marked with * are from [11]. Methods marked with † use semantic masks precomputed offline.

| Method | ATE RMSE (cm) |
|---|---|
| ORB-SLAM2 | 1.0 |
| **Semantic Geometric SLAM** | |
| RDS-SLAM (SegNet) | 0.4 |
| Dynamic-VINS | 0.7 |
| RDS-SLAM (Mask-RCNN) | 0.8 |
| VDO-SLAM† | 0.9 |
| Panoptic-SLAM | 1.8 |
| **Semantic NeRF SLAM** | |
| NIS-SLAM* | 2.1 |
| vMAP* | 2.4 |

**Table 4**
Performance comparison on the Replica dataset. Methods marked with † use semantic masks or labels precomputed offline. Values marked with * are from [11]. The average trajectory accuracy (ATE RMSE in centimeters) and semantic reconstruction quality (mIoU percentage) across all sequences are presented. Methods are ordered by increasing semantic reconstruction quality.

| Method | ATE RMSE (cm) | mIoU (%) |
|---|---|---|
| **Semantic NeRF SLAM** | | |
| NIDS-SLAM* | 0.80 | 82.37 |
| SNI-SLAM* | 0.46 | 83.62 |
| DNS-SLAM* | 0.45 | 84.77 |
| **Semantic GS SLAM** | | |
| NEDS-SLAM* | 0.35 | 90.78 |
| SGS-SLAM† | 0.41 | 92.72 |
| SemGauss-SLAM* | 0.33 | 93.71 |
| GS3LAM† | 0.58 | 96.63 |

### *4.3. Results*

In the initial phase of our comparative study, we evaluate Semantic SLAM systems by assessing their accuracy in pose estimation and trajectory computation under challenging conditions, namely dynamic environments.

Table 3 compares the accuracy of the selected Semantic SLAM systems against the Geometric SLAM approach ORB-SLAM2, using the TUM RGB-D dataset. All Semantic Geometric SLAM systems, except Panoptic SLAM, demonstrate that incorporating semantic data enhances the quality of pose estimation. RDS-SLAM with SegNet achieves the best performance with an ATE RMSE of 0.4 cm. Dynamic-VINS shows excellent performance with 0.7 cm error, while RDS-SLAM [Mask-RCNN] and VDO-SLAM also show slightly better results, with improvements of 0.8 cm and 0.9 cm respectively. For comparison, we include state-of-the-art results for two NeRF-based Semantic SLAM systems, NIS-SLAM and vMAP, although their execution was not possible on the NVIDIA Jetson AGX Orin due to insufficient resources. We observe that, at present, they achieve comparable accuracy, but it is still lower than that of geometric SLAM. This remains the case even though vMAP is built on ORB-SLAM3, using only its tracking component. The evaluation of both selected Semantic GS SLAM systems (GS3LAM and SGS-SLAM) could not be performed on the TUM RGB-D dataset because it lacks the necessary semantic annotations. While this dataset provides accurate ground truth trajectories and depth information, it lacks the per-frame semantic labels that are fundamental to these approaches. This limitation underscores a broader challenge in evaluating semantic SLAM systems on traditional benchmarks, which were primarily designed for geometric reconstruction and pose estimation tasks.

To evaluate the Semantic NeRF-based SLAM and GS-based SLAM systems, the Replica dataset was used, and the results are presented in Table 4. As we were unable to conduct our own tests on some systems due to insufficient resources, we present comparative results reported in the state-of-the-art. The consistent sub-centimeter accuracy achieved by the majority of these Semantic SLAM systems highlights their maturity and reliability in structured indoor environments. SemGauss-SLAM leads with an ATE RMSE of 0.33 cm, closely followed by NEDS-SLAM at 0.35 cm. SGS-SLAM achieves 0.41 cm, while GS3LAM shows a slightly higher error at 0.58 cm. Notably, SGS-SLAM incorporates loop closure, which contributes to its robust performance. GS3LAM achieves the highest reported mIoU at 96.63%, followed by SemGauss-SLAM at 93.71% and SGS-SLAM at 92.72%. However, GS3LAM and SGS-SLAM consume semantic inputs prepared offline. Their mIoU values therefore characterize semantic fusion and reconstruction given these inputs, rather than the performance of a complete online semantic perception and SLAM pipeline. Even lower performing methods maintain mIoU scores above 82%, indicating robust semantic understanding between different SLAM architectures. GS3LAM stands out for its strong semantic reconstruction capabilities, while the architecture of SGS-SLAM enables trajectory correction, making both methods valuable for structured indoor environments. However, these accuracy and semantic-quality results must be interpreted together with embedded resource measurements, because high mIoU does not necessarily imply feasible online deployment on a constrained platform.

The resource utilization results in Table 6 and Fig. 9 must be interpreted according to the scope of the executed pipeline. Among the evaluated methods whose measurements include online semantic processing, Dynamic-VINS provides the best balance between accuracy and efficiency, reaching 24.9 FPS with 8.29 GB of RAM and a power consumption of 12 W. VDO-SLAM reaches a measured SLAM-loop throughput of 20.9 FPS with 6.62 GB of RAM and a power consumption of 10.9 W. However, these measurements exclude Mask R-CNN inference because the semantic masks are generated offline. The result is therefore not directly comparable with the end-to-end throughput of systems whose semantic processing is executed online. The same qualification applies to GS3LAM and SGS-SLAM. Their measured throughput of 0.013 to 0.014 FPS includes Gaussian-based tracking, mapping, semantic fusion, rasterization, and map optimization, but excludes the generation of the semantic masks or labels. Although both systems remain far from real time even under this favorable setting, their reported throughput remains a lower bound on the computational cost of a complete end-to-end Semantic SLAM pipeline. The choice of segmentation method also significantly impacts overall performance, as shown by the segmentation timing bars in Fig. 9. The different variants of RDS-SLAM highlight this distinction clearly: RDS-SLAM with SegNet outperforms Mask-RCNN by delivering more efficient segmentation, while still maintaining competitive accuracy. This performance gap is due to their different neural network architectures, with SegNet specifically designed for real-time applications. Among the geometric SLAM systems, Panoptic SLAM barely achieves 2.7 FPS, which indicates that panoptic semantics are not yet efficient enough for real-time embedded performance.

**Computational bottlenecks of neural scene representations.** The poor embedded behavior of neural and Gaussian scene representations on the Jetson AGX Orin is explained by the structure of their online optimization loop. In NeRF-based SLAM, tracking and mapping require repeated volumetric rendering from selected pixels or keyframes. For each ray, many samples are evaluated by an MLP or hybrid decoder to predict density, color, and semantic features. This creates a large number of small neural queries, followed by backpropagation through photometric, depth, pose, and semantic losses. The cost scales with the number of sampled rays, the number of samples per ray, the size of the decoder, and the number of optimization iterations per frame or keyframe. Because these steps must be repeated while the robot is moving, NeRF-based methods are limited by online MLP training, ray sampling, activation memory, and memory-bandwidth pressure rather than by final rendering quality alone. On embedded platforms, this is particularly restrictive because the GPU must share power and memory bandwidth with CPU-side tracking, data loading, and semantic preprocessing.

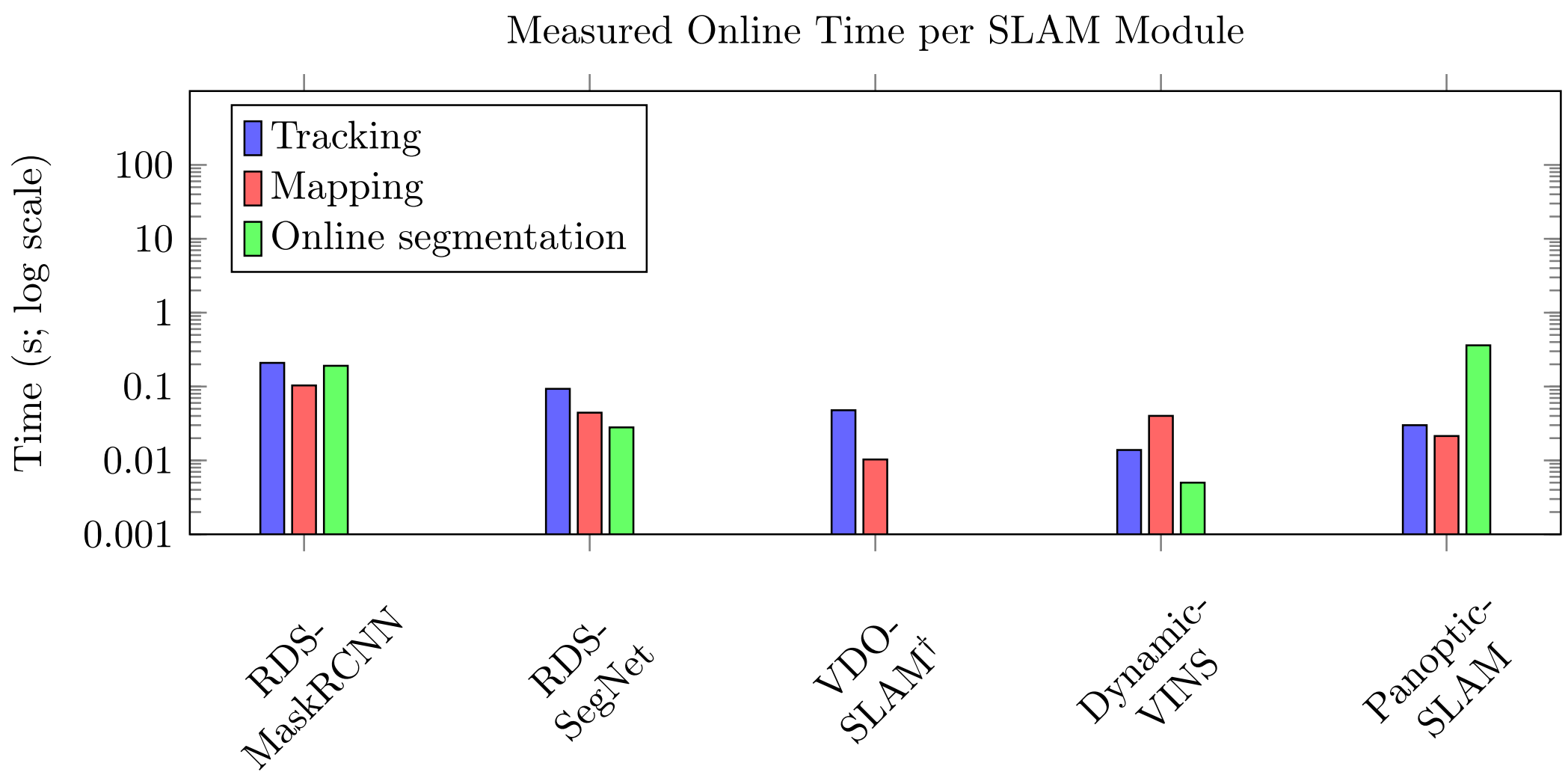

**Fig. 9.** Measured per-module latency of Semantic Geometric SLAM approaches on the NVIDIA Jetson AGX Orin. Processing times are reported in seconds using a logarithmic scale and are decomposed into tracking (blue), mapping (red), and online semantic processing (green). For VDO-SLAM, instance segmentation is performed offline and excluded from the measured runtime. VDO-SLAM is therefore marked with †, and the absence of a green bar does not represent zero semantic-processing cost.

In 3D Gaussian Splatting SLAM, the bottleneck is different but still severe. Rendering is faster than NeRF because the representation is explicit, but real-time SLAM requires repeated differentiable rasterization and backpropagation while the Gaussian set is still growing. In our GS3LAM run, the map grows from about 0.815 million Gaussians at the first frame to 4.515 million Gaussians at the end of the sequence. The serialized parameter file is 499.7 MiB, and the semantic feature representation alone occupies 275.6 MiB ($4{,}515{,}403 \times 16$ float32 values). This is only the stored semantic descriptor; during optimization, gradients and optimizer states further increase the active memory footprint. GS3LAM performs pose tracking by repeatedly rendering the full Gaussian map and backpropagating photometric, depth, and semantic losses to the camera pose. Mapping is heavier: each frame first renders the current map to identify uncovered regions, inserts new Gaussians from RGB-D observations, rebuilds optimizer state over millions of primitives, and then optimizes the Gaussian parameters over the current frame and sampled keyframes. The final renderer reaches 7.67 FPS in evaluation, but the online SLAM loop reaches only 0.013 FPS because it is dominated by optimization rather than by feed-forward rendering. The difference between the logged mapping-iteration time (534 ms) and the full mapping-frame latency (76.42 s) also indicates that iteration timers that do not synchronize all GPU work can underestimate the effective embedded latency.

SGS-SLAM exhibits the same class of bottleneck with a different semantic design. Its final map contains 4.808 million Gaussians after pruning, with a 298.2 MiB parameter file and separate 312 MiB RGB and semantic PLY exports. Unlike GS3LAM, SGS-SLAM stores semantic colors rather than a 16-dimensional semantic feature vector, which reduces the serialized semantic tensor size. However, when semantic data are loaded, each optimization iteration performs three differentiable rendering passes before the backward pass: RGB rendering, depth/silhouette rendering, and semantic-color rendering. The method therefore pays for repeated full-map rasterization, semantic loss computation, optimizer-state updates, pruning, and keyframe-window optimization at every frame. The measured averages are 1.09 s per tracking iteration and 1.17 s per mapping iteration, which become 43.54 s and 70.05 s per frame with 40 tracking and 60 mapping iterations. These measurements show that the main embedded bottlenecks are not only low FPS, but also unbounded Gaussian-map growth, per-primitive semantic storage, repeated full-map rasterization, optimizer state traffic, and CUDA synchronization.

Table 5 summarizes the measured factors behind this conclusion. Making GS-based Semantic SLAM suitable for embedded deployment will therefore require substantial redesign, such as bounded map growth, fewer or compressed semantic channels, reduced optimization frequency, asynchronous mapping, quantized parameters, and hardware-aware rasterization kernels.

### 4.4. Discussion

Semantic SLAM still faces major embedded-deployment challenges. These include the cost of semantic inference, map growth, and irregular online optimization. We organize the following outlook around three directions. Resource-bounded perception and mapping control inference cost and representation growth. Uncertainty-aware resource allocation determines when costly processing or adaptation is needed. Joint hardware and algorithm design assigns each task to the most suitable processing unit. Together, these directions target bounded memory usage, predictable execution time, and limited energy consumption.

**Deployment Implications of Explicit and Hybrid Neural Representations.** Our measurements in Tables 5 and 6 lead to a clear deployment conclusion: in their evaluated forms, GS3LAM and SGS-SLAM [26,28] are not viable for real-time operation on memory-constrained embedded platforms. Even with semantic masks or labels computed offline, they use 16.10 to 16.56 GB of RAM and reach only 0.013 to 0.014 FPS. Their final serialized maps occupy 298.2 to 499.7 MiB, but this on-disk size does not represent the full active working memory required during online map optimization, which additionally includes gradients, optimizer states, visibility and rasterization buffers, and temporary tensors. Reducing the final map file alone is therefore insufficient to enable embedded deployment.

This behavior is partly explained by the explicit nature of 3D Gaussian Splatting: each primitive stores position, scale, rotation, opacity, appearance, and, when applicable, semantic attributes [26,28,185]. Consequently, both map storage and optimization state grow with

**Table 5**
Detailed bottlenecks observed for GS-based Semantic SLAM on the NVIDIA Jetson AGX Orin. Both evaluated methods rely on semantic masks precomputed offline; semantic inference is therefore excluded from all latency and throughput measurements.

| Observed factor | GS3LAM† | SGS-SLAM† |
|---|---|---|
| Semantic input | Precomputed offline | Precomputed offline |
| Semantic inference | Excluded | Excluded |
| Final map size | 4.515M Gaussians | 4.808M Gaussians |
| Serialized map size | 499.7 MiB | 298.2 MiB |
| Semantic representation storage | 275.6 MiB | 59.6 MiB |
| Tracking optimization | 40 iterations/frame | 40 iterations/frame |
| Mapping optimization | 60 iterations/frame | 60 iterations/frame |
| Tracking latency | 47.99 s/frame | 43.54 s/frame |
| Mapping latency | 76.42 s/frame | 70.05 s/frame |
| SLAM-loop throughput | 0.013 FPS | 0.014 FPS |
| Main bottleneck | Semantic feature storage | Multi-pass semantic rendering |

**Table 6**
Resource utilization comparison on the NVIDIA Jetson AGX Orin. Values represent RAM usage (in GB), power consumption (in W), and processing speed (in frames per second).

| Method | RAM (GB) | Power (W) | FPS |
|---|---|---|---|
| **Semantic Geometric SLAM** | | | |
| Panoptic-SLAM | 10.03 | 16.1 | 2.7 |
| RDS-SLAM (Mask-RCNN) | 9.46 | 16.0 | 4.8 |
| RDS-SLAM (SegNet) | 8.63 | 14.5 | 10.7 |
| VDO-SLAM† | 6.62 | 10.9 | 20.9 |
| Dynamic-VINS | 8.29 | 12.0 | 24.9 |
| **Semantic GS SLAM** | | | |
| GS3LAM† | 16.56 | 15.1 | 0.013 |
| SGS-SLAM† | 16.10 | 17.6 | 0.014 |

Methods marked with † use semantic masks or labels precomputed offline. Semantic inference is excluded from the reported runtime, memory, and power measurements. Their FPS values therefore characterize the executed SLAM loop rather than a complete end-to-end Semantic SLAM pipeline.

the number of Gaussians. Without aggressive pruning, compression, or bounded submapping, uncompressed GS maps therefore tend to be larger than compact implicit or hybrid neural-field representations for comparable scene coverage. This is a tendency rather than a universal ordering, because compressed GS representations and high-resolution grid-based neural fields can change the trade-off. The original NeRF used a positionally encoded coordinate MLP that was compact on disk but expensive to train and query [184]. Many recent neural implicit SLAM methods instead combine learned spatial encodings, such as multi-resolution hash grids, feature planes, or tetrahedral features, with shallow MLP decoders [131,134,143]. These hybrid designs reduce decoder size and offer a path toward smaller serialized representations and lower working-memory pressure than the evaluated GS maps, provided that the learned encoding remains bounded. Their actual memory advantage depends on the type, resolution, and size of that encoding; ray sampling, differentiable rendering, and repeated online backpropagation remain major bottlenecks. Further work is therefore required to improve rendering and online-optimization speed while enforcing bounded memory consumption.

**Resource-Bounded Semantic Perception and Mapping.** Embedded Semantic SLAM should combine lightweight or quantized semantic models with resource-bounded map representations. When appropriate, semantic inference may be executed selectively, for example when scene changes, tracking residuals, or semantic uncertainty indicate that an update is required, while labels or features are propagated between successive updates. The mapping module should maintain a bounded active region, archive inactive submaps, and prune or merge redundant points, voxels, neural-field parameters, or Gaussians.

**Uncertainty-Aware Resource Allocation and Adaptation.** Lightweight uncertainty indicators, such as semantic confidence, geometric residuals, pose covariance, or temporal inconsistencies, can be used to control the execution of costly SLAM modules. They may trigger semantic inference, keyframe insertion, or local map refinement, while confident and redundant observations can be processed with reduced computational effort. When online adaptation is required, the main backbone can remain frozen while only lightweight adapters, classification heads, or semantic prototypes are updated using a bounded replay buffer. Such updates should run as low-priority tasks under fixed latency, memory, and power budgets.

**Joint Hardware and Algorithm Design for Semantic SLAM.** Existing accelerators mainly target regular and highly parallel kernels, including feature extraction, optical flow, neural-network inference, and matrix operations, whereas irregular data association, map management, and nonlinear optimization generally remain on programmable processors. Consequently, co-design should combine algorithmic adaptation with heterogeneous execution rather than merely offloading isolated kernels. Semantic inference can be executed selectively or at a lower rate, for example when triggered by scene changes, tracking degradation, or uncertainty. Its regular neural-network operators can be mapped onto a GPU, DLA, or FPGA, while tracking and nonlinear optimization remain on the CPU. Shared visual features, asynchronous execution, and zero-copy buffers can reduce duplicated computation, synchronization, and data transfers. Finally, a deadline- and power-aware scheduler can prioritize tracking and adapt semantic inference and map refinement to the available computational and energy budgets.

## 5. Conclusion

This survey presents a comparative review of state-of-the-art Semantic SLAM systems from an embedded-deployment perspective. To our knowledge, this is the first survey whose primary scope is the feasibility of Semantic SLAM on embedded systems, rather than a general taxonomy of semantic, implicit, NeRF-, or GS-based SLAM. Through experimental analysis, three semantic-aware architectural approaches: Geometric-based, NeRF-based, and 3D GS-based SLAM systems, were examined. In addition, our study evaluates the feasibility of deploying these systems on platforms with limited resources, with a particular focus on the NVIDIA Jetson AGX Orin, a representative embedded system commonly used in robotics applications. The Jetson experiments lead to three main conclusions. First, Semantic Geometric SLAM systems currently provide the most viable solution for real-time embedded deployment. Dynamic-VINS reaches 24.9 FPS with online semantic processing, whereas VDO-SLAM†reaches 20.9 FPS for the measured SLAM loop only, excluding its offline Mask R-CNN inference. Second, the apparent efficiency of several methods depends strongly on whether semantic information is computed online or preprocessed offline, which must be reported when evaluating embedded readiness. Third, GS-based semantic SLAM can achieve strong semantic reconstruction quality, but the tested systems still require more than 16 GB of RAM and more than 15 W while running at only 0.013 to 0.014 FPS. The underlying bottlenecks are online MLP training and ray sampling for NeRF-based SLAM, and differentiable rasterization, densification, semantic-feature storage, and Gaussian-parameter optimization for GS-based SLAM. These results show that memory optimization, power-aware semantic processing, and effective management of semantic features remain key requirements for deployment on resource-constrained platforms.

As SLAM continues to evolve, addressing several challenges will be crucial to improving the robustness, scalability, and real-time capabilities of SLAM systems, particularly for deployment in real-world applications. Future embedded Semantic SLAM research should therefore focus on three coupled directions: resource-bounded semantic perception and mapping, uncertainty-aware resource allocation and lightweight adaptation, and joint hardware and algorithm design. Together, these directions provide concrete implementation routes for controlling memory growth, execution latency, and energy consumption. With ongoing advancements in these areas, the next generation of SLAM systems will be more adaptable, efficient, and capable of handling complex, real-world scenarios.

## CRediT authorship contribution statement

**Calvin Galagain:** Writing – review & editing, Writing – original draft, Visualization, Software, Methodology, Investigation, Data curation, Conceptualization. **Martyna Poreba:** Writing – review & editing, Writing – original draft, Validation, Supervision, Project administration, Funding acquisition, Conceptualization. **François Goulette:** Writing – review & editing, Validation, Supervision, Project administration, Methodology, Conceptualization.

## Declaration of competing interest

The authors declare that they have no known competing financial interests or personal relationships that could have appeared to influence the work reported in this paper.

## Acknowledgments

This work was partially funded by the French Defence Innovation Agency (Agence de l'innovation de défense, AID).

## Data availability

Data will be made available on request.

**Calvin Galagain** is a Ph.D. student working jointly between CEA-List and ENSTA Paris. In 2023, he earned his engineering degree from ESIEE Paris and a Master's in Mathematics, Vision, and Artificial Intelligence (MVA) from ENS Paris-Saclay. During his internship, he gained experience in segmenting point clouds captured by a LiDAR-based SLAM system. His current research focuses on semantic SLAM systems, with a particular emphasis on optimizing these systems for platforms with limited resources.

**Martyna Poreba** is a research engineer at CEA-List, specializing in multimodal perception for embedded systems. She holds a Ph.D. in Computer Science from MINES Paris, where her early work focused on mobile mapping, photogrammetry, and 3D data processing. She then conducted research on cross-domain and multi-source image retrieval at the French Ministry of Ecology. Currently, her research focuses on optimizing machine learning models for limited resources devices, aiming to advance AI applications in real-world scenarios, including object detection and tracking, as well as visual–inertial navigation systems (VINS) enhanced with semantic data.

**François Goulette** graduated from Mines Paris with an engineer degree in 1992, and a Ph.D. in computer science and robotics in 1997. He obtained an Habilitation to Conduct Research from Sorbonne University in 2009. He worked as a research engineer at EDF and later as professor at Mines Paris - PSL University. Since 2022, he is full Professor and deputy director of the U2IS (Robotics and AI) Lab at ENSTA Paris - Institut Polytechnique de Paris. His research focuses on 3D perception, mobile mapping, photogrammetry, and 3D data processing for robotics and autonomous vehicles.